\documentclass[11pt]{article}

\usepackage{jmlr2e}
\usepackage{amsmath,rotating}

\newcommand{\tpr}{\mathop{\mathit{tpr}}}
\newcommand{\fpr}{\mathop{\mathit{fpr}}}
\renewcommand{\prec}{\mathop{\mathit{prec}}}

\ShortHeadings{Population \& Empirical PR Curves to Assess Ranking}{Hughes-Oliver}
\firstpageno{1}

\begin{document}
	
\title{Population and Empirical PR Curves for Assessment of Ranking Algorithms}
	
\author{\name Jacqueline M. Hughes-Oliver \email hughesol@ncsu.edu \\
		\addr Department of Statistics\\
		North Carolina State University\\
		Raleigh, NC 27695-8203, USA}

\editor{TBD}

\maketitle

\begin{abstract}
The ROC curve is widely used to assess the quality of prediction/classification/ranking algorithms, and its properties have been extensively studied. The precision-recall (PR) curve has become the de facto replacement for the ROC curve in the presence of imbalance, namely where one class is far more likely than the other class. While the PR and ROC curves tend to be used interchangeably, they have some very different properties. Properties of the PR curve are the focus of this paper. We consider: (1) population PR curves, where complete distributional assumptions are specified for scores from both classes; and (2) empirical estimators of the PR curve, where we observe scores and no distributional assumptions are made. The properties have direct consequence on how the PR curve should, and should not, be used. For example, the empirical PR curve is not consistent when scores in the class of primary interest come from discrete distributions. On the other hand, a normal approximation can fit quite well for points on the empirical PR curve from continuously-defined scores, but convergence can be heavily influenced by the distributional setting, the amount of imbalance, and the point of interest on the PR curve.
\end{abstract}


\section{Introduction}
ROC curves provide concise and informative summaries of the effectiveness of prediction or classification or ranking algorithms for distinguishing between two classes~\citep[see, for example,][]{Pepe2003,Fawcett2006,KrzanowskiHand2009}. For this article, we refer to these classes as positive ($+$) and negative $(-)$. The vertical axis displays the {\em true positive rate}, $\tpr$, which is the probability of predicting a $+$ as $+$, while the horizontal axis displays the {\em false positive rate}, $\fpr$, which is the probability of predicting a $-$ as $+$. The curve arises from varying the threshold applied to {\em ranking-algorithm scores} in order to make predictions for individual instances; instances with scores larger than the threshold are predicted as $+$, and otherwise as $-$. As the threshold changes, $\tpr$ and $\fpr$ may also change. Clearly, the ideal point on the ROC curve is the pair (0,1), where there are no false positives and all members of the $+$ class are identified as such. But, akin to standard hypothesis testing where one must acknowledge that errors will occur, the task is to find an appropriate balance between the $\fpr$ being small and the $\tpr$ being large.

If we emulate hypothesis testing to limit the probability of a type I error to $\alpha$ while maximizing the power, this equates to setting $\fpr\le\alpha$ then finding the largest $\tpr$. The ROC curve is quite often strictly increasing, and so this largest $\tpr$ usually occurs at the vertical axis value that corresponds to horizontal axis value of $\alpha$. Indeed, for a fixed $\fpr=\alpha$, the ROC point $(\alpha,\tpr)$ may be cast as a most powerful Neyman Pearson test where the null hypothesis is that the instance is $-$ and the alternative hypothesis is that the instance is $+$. The difficulty comes in two ways: (1) it is often not clear what constitutes an appropriate value for $\alpha$, or the maximum allowed $\fpr$; and (2) once $\alpha$ is specified, data is used to estimate the threshold (unlike usual hypothesis testing, where the threshold corresponding to $\alpha$ is clearly defined) at which $\tpr$ is determined, thus extra uncertainty is injected into the process. Threshold determination is important because a ranking algorithm could perfectly separate the classes yet still not give perfect results if a poor choice is made regarding a threshold \citep{Fawcett2006}.

When comparing two ROC curves (from two ranking algorithms), the ideal scenario is that one curve dominates the other. More specifically, for each fixed value of $\fpr$, $\tpr_A\ge \tpr_B$, i.e., the $\tpr$ from curve A is at least as large as the $\tpr$ from curve B. Alternatively, dominance could be described as $\fpr_A\leq \fpr_B$ for each fixed value of $\tpr$. In this case, ranking algorithm A is uniformly better than ranking algorithm B. In practice, however, we more often observe ROC curves that cross. Consider the two ROC curves shown in Figure~\ref{fig:caseAB}(b) (full explanation to follow in Section~\ref{sec:popncurves}). Ranking algorithm A is better if we consider $\fpr$ values above 0.2 tolerable, while ranking algorithm B is better if we need $\fpr$ less than 0.2. So we are faced with the question of which algorithm is best, subject to a maximum tolerable value for $\fpr$. This translates into how to choose the threshold.

A natural way to provide greater focus to small values of $\fpr$ is afforded by the increasingly popular precision-recall (PR) curve \citep{RaghavanBollmannJung1989,Provostetal1998,DavisGoadrich2006,Brodersenetal2010}. While the PR curve is used quite often in machine learning and information retrieval, it has received relatively little attention regarding its statistical inferential properties; the most notable exceptions are~\citet{DavisGoadrich2006}, \citet{ClemenconVayatis2009}, and \citet{Boydetal2012}. The PR curve parallels the ROC curve in many ways, but can actually be more informative in that it is affected by what is commonly called the prior probabilities. Prior probabilities represent class membership, $\pi_+$ and $\pi_-$, and are rarely known in practice, but something can usually be said about the so-called {\em skew} defined as $\pi_-/\pi_+$. ROC curves provide no information about {\em skew} and in fact are invariant to {\em skew} \citep{Pepe2003,Fawcett2006,KrzanowskiHand2009}. On the other hand, PR curves are very much a function of the class probabilities and tend to be more useful in cases of imbalance (i.e., skew not equal to one); see \citet{RaghavanBollmannJung1989,DavisGoadrich2006,ClemenconVayatis2009,Boydetal2012}. 

For example, consider the extremely desirable ROC point $(\fpr,\tpr)=(.01,.99)$ when $\pi_+=.001$. If the goal is to rank-order instances to find members of the $+$ class (e.g., predicting emails as scam), one will be interested in the probability that an instance is $+$ given it was predicted as $+$. This probability is a minuscule .09, thus suggesting the desirable ROC point is actually a complete failure. The PR curve places this ROC point $(.01,.99)$ when $\pi_+=.001$ in an undesirable region because it becomes the PR point $(.99,.09)$, where optimality is far away at $(1,1)$. Using exactly the same input as the ROC curve, the PR curve provides a summary regarding utility of an algorithm for finding members of the $+$ class, making the PR curve a very viable tool. Proper use of the PR curve (and any summaries obtained from it) should be guided by the properties and limitations of this curve.

The remainder of this paper is organized as follows. Section~2 defines and presents properties of the PR curve when complete distributional assumptions are specified for scores  from both classes. Complete proofs are given for properties of these population PR curves. Six sets of distributional assumptions serve to illustrate the various properties, including bi-normal, bi-beta, overlapping uniforms, subset ranges for continuously-defined scores, and overlapping ranges for discretely defined scores. Section~3 defines and present properties of a nonparametric (empirical) estimator of the PR curve; small-sample and asymptotic behavior are studied. Concluding remarks are given in Section~4.

\section{Defining PR Curves from Population Scores}
\label{sec:popncurves}
In this section, we first define population PR curves as arising when full distributional information is known, then investigate some relevant properties. Finally, full details of these population PR curves and their properties are illustrated using six cases that include discrete and continuous scores.

\subsection{Definitions}
Ranking algorithms produce a {\em score}, $S$, that is used in predicting class membership for an instance (also commonly referred to as an item, record, object, etc.). Without loss of generality, assume large values of $S$ are consistent with membership in class $+$ and small values of $S$ are consistent with membership in class $-$. $S$ is viewed as a random variable whose distribution depends on the true (unknown) class. For a specified threshold $t$, one can consider the joint distribution of true and predicted class membership as given in Table~\ref{tab:confmat}. This {\em population-level confusion matrix} can change with changing values of threshold $t$, but for a given $t$ it can be summarized using only three numbers. In fact, provided $0<\pi_+<1$ and $0<\Pr(S>t)$, the confusion matrix can be uniquely and equivalently represented using either triplet
\begin{equation} \textstyle\tpr_t=\Pr(S>t|+) \qquad \& \qquad \fpr_t=\Pr(S>t|-) \qquad \& \qquad \pi_+ 
\label{eq:cf-to-roc}
\end{equation}
or triplet
\begin{equation} \textstyle \tpr_t  \qquad \& \qquad \prec_t=\Pr(+|S>t)  \qquad \& \qquad \pi_+ 
\label{eq:cf-to-pr}
\end{equation}
See~\citet{Pepe2003} and~\citet{DavisGoadrich2006} for details.
\begin{table}
\begin{center}
	\begin{tabular}{rc|cc|l}
		\multicolumn{2}{c}{} & \multicolumn{2}{c}{Truth} \\
		\multicolumn{2}{c}{} & $+$ & \multicolumn{1}{c}{$-$} \\ \cline{3-4}
		Predicted & $+$ & $\Pr(S>t,+)$ & $\Pr(S>t,-)$ & $\Pr(S>t)$ \\
		& $-$ & $\Pr(S\leq t,+)$ & $\Pr(S\leq t,-)$ & $\Pr(S\leq t)$ \\ \cline{3-4}
		\multicolumn{2}{c}{} & $\pi_+$ & \multicolumn{1}{c}{$\pi_-$}	\end{tabular}
\end{center}
	\caption{{\bf Population-Level Confusion Matrix}, namely the joint distribution of true and predicted class membership \label{tab:confmat}}
\end{table}

The ROC curve is based on representation~(\ref{eq:cf-to-roc}) and plots $\tpr_t$ as a function of $\fpr_t$. For all values of $t$, the ROC curve plots $(x_{\mathrm{ROC}},y_{\mathrm{ROC}})$, where
\begin{equation*}
\begin{array}{rcccccl}
x_{\mathrm{ROC}} & = & \fpr_t & = & \Pr(S>t|-) &=& 1-F_-(t) \\
y_{\mathrm{ROC}} & = & \tpr_t & = & \Pr(S>t|+) &=& 1-F_+(t) 
\end{array},
\end{equation*}
and $F_-(\cdot)$ and $F_+(\cdot)$ are the distribution functions of score $S$ for different classes. To more directly identify $y_{\mathrm{ROC}}$ as a function of $x_{\mathrm{ROC}}$ rather than as a function of $t$, we can write $t=F_-^{-1}(1-x_{\mathrm{ROC}})$, where $F_-^{-1}(\cdot)$ is the generalized inverse of distribution function $F_-(\cdot)$. Hence, the ROC curve is 
\begin{equation}
y_{\mathrm{ROC}}(x_{\mathrm{ROC}}) = 1-F_+\left( F_-^{-1}(1-x_{\mathrm{ROC}}) \right), \qquad 0 < x_{\mathrm{ROC}} < 1.
\label{eq:roccdfs}
\end{equation}
It is interesting to note that although motivated by triplet representation~(\ref{eq:cf-to-roc}) of the population-level confusion matrix in Table~\ref{tab:confmat}, the ROC curve actually ignores class probability $\pi_+$ that is a part of the triplet. This is often described as an advantage of the ROC curve \citep{Pepe2003,Fawcett2006,KrzanowskiHand2009}. But, depending on the intended usage of ranking results, this could be a disadvantage, as demonstrated in the introduction.

The PR curve is based on representation~(\ref{eq:cf-to-pr}) and plots $\prec_t$ as a function of $\tpr_t$ for all values of $t$. The {\em precision}, $\prec$, is the probability that an instance is $+$ given it was predicted $+$. In medical testing, precision is usually referred to as the {\em positive predictive value}. Originating in the specialty of information retrieval, the term {\em recall} is equivalent to $\tpr$, hence the name of the PR curve. For all values of $t$, the PR curve plots
$(x_{\mathrm{PR}},y_{\mathrm{PR}})$, where
\begin{equation}
\begin{array}{rclcccl}
x_{\mathrm{PR}} & = & \tpr_t & = & \Pr(S>t|+) &=& y_{\mathrm{ROC}} \\
y_{\mathrm{PR}} & = & \prec_t & = & \Pr(+|S>t) &=& \frac{\pi_+}{\pi_+ + \pi_- \left(x_{\mathrm{ROC}} / y_{\mathrm{ROC}}\right) } 
\end{array},
\label{eq:pr-from-roc}
\end{equation}
and hence
\begin{equation}
y_{\mathrm{PR}}(x_{\mathrm{PR}}) = 
\frac{\pi_+}{\pi_+ + \pi_- \left[ 1-F_-\left( F_+^{-1}(1-x_{\mathrm{PR}}) \right) \right]/x_{\mathrm{PR}} }, \qquad 0 < x_{\mathrm{PR}} < 1.
\label{eq:prcdfs}
\end{equation}

\subsection{Properties}
\label{sec:properties}
We now consider some useful properties of PR curves, with proofs given in the Appendix. 
Some, but not all, of these properties have been presented or even proved elsewhere, while for others we provide refined statements. \citet{BoydEngPage2013,BoydEngPage2013error} suggest that the PR curve {\it decreases} to $\pi_+$ as recall increases to one; while this is often the case, we demonstrate that this is not guaranteed. \citet{ClemenconVayatis2009} correctly argue that the PR curve {\it approaches} $\pi_+$ as recall increases to one when $F_+$ and $F_-$ have the same support; we derive the limit even when $F_+$ and $F_-$ have differing support. In fact, \citet{ClemenconVayatis2009} limit all discussion to continuous $F_+$ and $F_-$ having the same support, but this paper takes a broader view on the set of possible $F_+$ and $F_-$ that might occur in practice. Although conditions for monotonicity of the PR curve have been addressed by \citet{ClemenconVayatis2009} and lower bounds on the PR curve have been addressed by \citet{Boydetal2012}, this paper adds to those contributions.

As above, the ranking-algorithm score $(S)$ is assumed to be a random variable with distribution functions $F_-(\cdot)$ and $F_+(\cdot)$ in the $-$ and $+$ classes. For each of $j=-$ and $j=+$, the possible values of the score range from $m_j$ to $M_j$. If scores are continuous random variables, they have densities $f_-(\cdot)$ and $f_+(\cdot)$.

Quantile functions or generalized inverse distribution functions are heavily used below. The usual definition is used, that is, $F^{-1}(p)=\inf\{z:F(z)\geq p\}$ for $0\leq p\leq 1$, where it is understood that $\inf\emptyset=\infty$. Many properties result~\citep[see~][]{EmbrechtsHofert2013}, including: $F^{-1}(p)$ is nondecreasing; if $F^{-1}(p)\in(-\infty,\infty)$, $F^{-1}(p)$ is left-continuous at $p$ and admits a limit from the right at $p$; $F(F^{-1}(p))\geq p$ and $F^{-1}(F(z))\leq z$; and $F(F^{-1}(p))= p$ and $F^{-1}(F(z))= z$ if $F(z)$ is strictly increasing. These properties allow us to conclude that $\lim_{p\uparrow1}F_j^{-1}(p)=F_j^{-1}(1)=M_j$ and $F_j^{-1}(0)=-\infty$, but $\lim_{p\downarrow0}F_j^{-1}(p)=m_j$.

Proofs for the following properties are provided in Appendix~A:
\begin{itemize}
	\item[P1.] The ROC curve is nondecreasing, with (one-sided) limiting values as follows:
	\begin{enumerate}
		\item[(a)] $\lim_{x_{\mathrm{ROC}}\downarrow 0}y_{\mathrm{ROC}}\left( x_{\mathrm{ROC}} \right)=1-F_+(M_-)$.
		\item[(b)] $\lim_{x_{\mathrm{ROC}}\uparrow 1}y_{\mathrm{ROC}}\left( x_{\mathrm{ROC}} \right)=1-F_+(m_-)$.
	\end{enumerate}
	
	\item[P2.] The PR curve is not necessarily monotone, with (one-sided) limiting values as follows:
	\begin{itemize}
		\item[(a)] $\lim_{x_{\mathrm{PR}}\uparrow1}y_{\mathrm{PR}}\left( x_{\mathrm{PR}} \right) = \frac{\pi_+}{\pi_+ + \pi_- [1-F_-(m_+)]}$.
		\item[(b)] If the maximum possible score among members of class $+$ is at least as large as the maximum possible score among members of class $-$ (i.e., $M_+\geq M_-$), then $\lim_{x_{\mathrm{PR}}\downarrow0}y_{\mathrm{PR}}\left( x_{\mathrm{PR}} \right) = \frac{\pi_+}{\pi_+ + \pi_- k}$, where $k=\lim_{t\uparrow M_+}\frac{f_-(t)}{f_+(t)}$ may be infinity.
		\item[(c)] If $M_+ < M_-$, then $\lim_{x_{\mathrm{PR}}\downarrow0}y_{\mathrm{PR}}\left( x_{\mathrm{PR}} \right) = 0$.
	\end{itemize} 
	
	\item[P3.] \underline{\em Monotonicity of the PR curve.}
	\begin{itemize}
		\item[(a)] If the ROC curve is concave and $M_+\geq M_-$, then the PR curve is nonincreasing.
		\item[(b)] If the ROC curve is convex, $M_+ < M_-$, and $\lim_{t\uparrow M_+} \left\{ \frac{f_-(t)}{f_+(t)}\left[1-F_+(t)\right] \right\}=0$, then the PR curve is nondecreasing.
	\end{itemize}
	
	\item[P4.] \underline{\em Chance Curves.} A ranking algorithm is useless if score distributions are the same across classes. If the populations of scores for $+$ and $-$ are identical, then chance ROC and PR curves are
	\begin{align*}  
	y_{\mathrm{ROC}}(x_{\mathrm{ROC}}) &\leq x_{\mathrm{ROC}} &  \qquad 0< x_{\mathrm{ROC}}< 1 \\
	y_{\mathrm{PR}}(x_{\mathrm{PR}}) &\geq \pi_+ &  \qquad 0< x_{\mathrm{PR}}< 1 ,
	\end{align*}
	with equality at values of $x_{\mathrm{ROC}}$ and $x_{\mathrm{PR}}$ for which $F_+(z)=1- x_{\mathrm{ROC}}$  and $F_+(z)=1- x_{\mathrm{PR}}$ for a unique $z$.
	
	\item[P5.] \underline{\em Perfect-Separation Curves.} The ideal ROC and PR curves occur when all scores for the $+$ class exceed all scores for the $-$ class, meaning $M_-<m_+$. The perfect-separation ROC and PR curves are actually not true functions because each curve can have multiple ordinates for the same abscissa. For this reason, it is more convenient to describe these perfect-separation curves as a function of threshold, as follows and graphed in Figure~\ref{fig:perfsep}:
	\[ \left( x_{\mathrm{ROC}}, y_{\mathrm{ROC}} \right) = \begin{cases}
	\left(0,1-F_+(t) \right) & M_+>t> m_+ \\
	\left(0,1 \right) & m_+\geq t>M_- \\
	\left(1-F_-(t),1 \right) & M_-\geq t>m_-
	\end{cases} \]
	and
	\[ \left( x_{\mathrm{PR}}, y_{\mathrm{PR}} \right) = \begin{cases}
	\left(1-F_+(t),1 \right) & M_+>t> m_+ \\
	\left(1,1 \right) & m_+\geq t>M_- \\
	\left(1,\frac{\pi_+}{\pi_+ + \pi_-[1-F_-(t)]} \right) & M_-\geq t>m_-
	\end{cases}. \]
	\begin{figure}
		\centering\includegraphics[width=.9\textwidth]{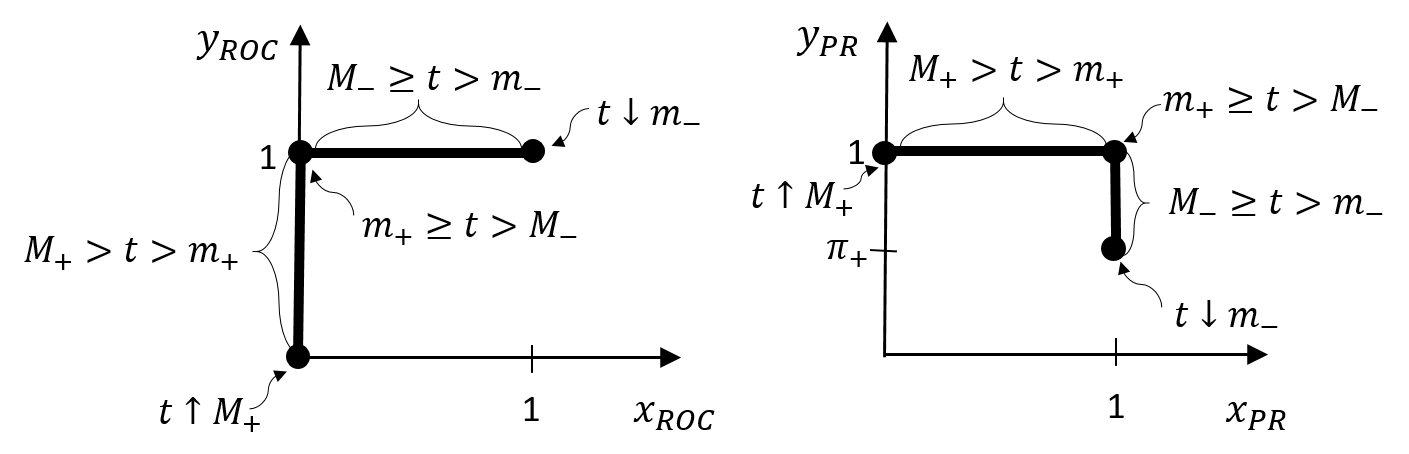}
		\caption{\label{fig:perfsep} Perfect-separation ROC and PR curves.}
	\end{figure}
	
	\item[P6.] \underline{\em Reverse-Separation Curves.} In the event that all scores for the $+$ class are exceeded by all scores for the $-$ class (i.e., $M_+<m_-$), one could simply multiply all scores by negative one to yield perfect-separation ROC and PR curves. Nonetheless, it is interesting to investigate the scenario when $M_+<m_-$ because it represents the lower bounds on these curves.  The reverse-separation ROC and PR curves are actually not true functions because each curve can have multiple ordinates for the same abscissa. For this reason, it is more convenient to describe these reverse-separation curves as a function of threshold, as follows and graphed in Figure~\ref{fig:revsep}:
	\[ \left( x_{\mathrm{ROC}}, y_{\mathrm{ROC}} \right) = \begin{cases}
	\left(1-F_-(t),0 \right) & M_->t>m_- \\
	\left(1,0 \right) & m_-\geq t>M_+ \\
	\left(1,1-F_+(t) \right) & M_+\geq t>m_+
	\end{cases} \]
	and
	\[ \left( x_{\mathrm{PR}}, y_{\mathrm{PR}} \right) = \begin{cases}
	\left(0,0 \right) & M_->t > M_+ \\
	\left(1-F_+(t),\frac{\pi_+[1-F_+(t)]}{\pi_+[1-F_+(t)] + \pi_-} \right) & M_+ \geq t >m_+
	\end{cases}. \]
	Moreover, the achievable lower bound curves are
	\[ y_{\mathrm{ROC}}(x_{\mathrm{ROC}}) \geq 0, \qquad 0<x_{\mathrm{ROC}}<1 \]
	and
	\[ y_{\mathrm{PR}}(x_{\mathrm{PR}}) \geq \frac{\pi_+\cdot x_{\mathrm{PR}}}{\pi_+\cdot x_{\mathrm{PR}} + \pi_-}, \qquad 0<x_{\mathrm{PR}}<1. \]
	\begin{figure}
		\centering\includegraphics[width=.78\textwidth]{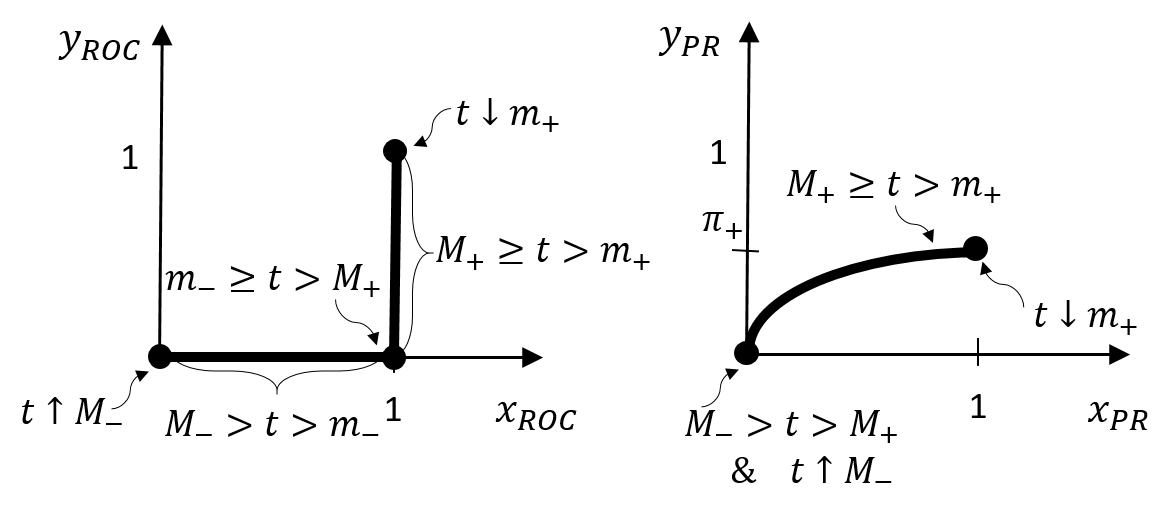}
		\caption{\label{fig:revsep} Reverse-separation ROC and PR curves.}
	\end{figure}
	
	\item[P7.] \underline{\em Invariance to increasing transformation.}
	ROC and PR curves are unaffected if the same increasing transformation is applied to scores for both classes.
	
\end{itemize}

\subsection{Illustrations}
\label{sec:illustrate}
\begin{figure}
	\includegraphics[width=\textwidth]{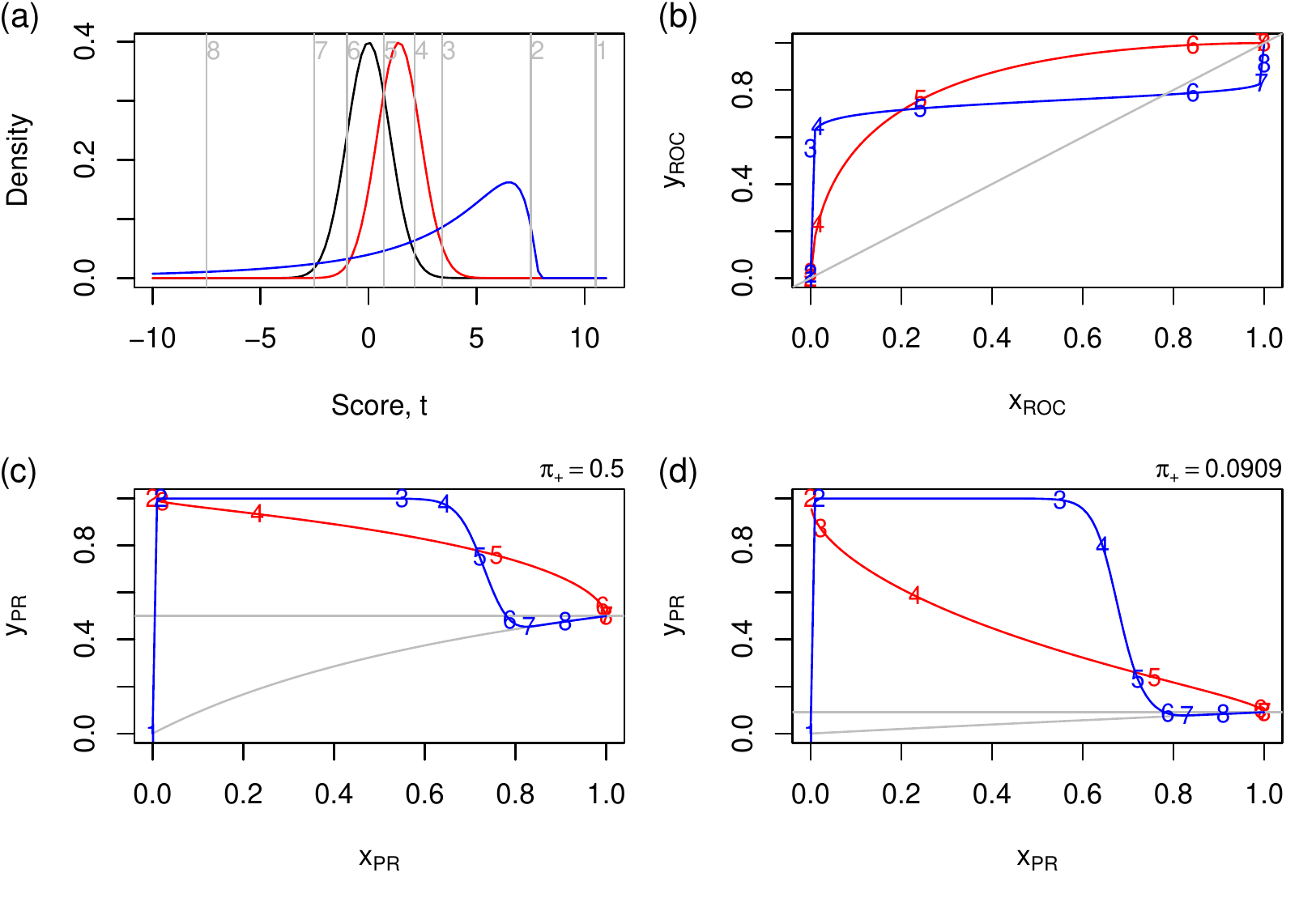}
	\caption{\label{fig:caseAB} Densities, ROC curves, and PR curves for Cases A and B. (a) Density functions for: the standard normal distribution as the black curve, corresponding to the $-$ class for both cases; the normal distribution with mean 1.4 and standard deviation 1 as the red curve, corresponding to the $+$ class for Case A; and the density corresponding to the $+$ class for Case B as the blue curve, obtained as $8-W$ where $W$ follows a lognormal distribution with parameters $\mu=1.4$ and $\sigma=1$. Eight thresholds are shown and labeled as vertical reference lines, and these thresholds are also shown in the ROC and PR curves. (b) ROC curves with labels corresponding to the thresholds displayed in (a): Case A is the red curve, Case B is the blue curve. The gray diagonal line is the chance curve. (c) PR curves with labels corresponding to the thresholds displayed in (a), assuming balanced classes so that $\pi_+=1/2$: Case A is the red curve, Case B is the blue curve. The gray horizontal line is the chance curve, while the other gray curve is the achievable lower bound curve. (d) Same as (c), except assuming the $-$ class is ten times as likely as the $+$ class so that $\pi_+=1/11$.}
\end{figure}

Properties P1--P7 are illustrated using six cases described below and depicted in Figures~\ref{fig:caseAB}--\ref{fig:caseF}. Cases~C and D are motivated by \citet{BoydEngPage2013}. Cases A and B are depicted in Figure~\ref{fig:caseAB}, Case C is depicted in Figure~\ref{fig:caseC}, Cases D and E are depicted in Figure~\ref{fig:caseDE}, and Case F is depicted in Figure~\ref{fig:caseF}. Each case may be viewed as output from a ranking algorithm, meaning scores are assigned to instances in the $-$ and $+$ classes. An effective ranking algorithm tends to produce highly separated scores for the classes.
\begin{description}
	\item[Case A:] Scores follow the popular bi-normal model, meaning scores from the $+$ class are distributed as normal with mean $\mu_+$ and variance $\sigma^2_+$, while scores from the $-$ class are distributed as normal with mean $\mu_-$ and variance $\sigma^2_-$. For consistency with large scores suggesting the $+$ class, $\mu_+>\mu_-$. For depiction in Figure~\ref{fig:caseAB}, values are set as $\mu_-=0,\ \mu_+=1.4$, and $\sigma^2_-=\sigma^2_+=1$. Because $m_-=m_+=-\infty$ and $M_-=M_+=+\infty$, the ROC curve starts at (0,0), is nondecreasing, and ends at (1,1), according to P1. The ROC curve is concave because $f_+(t)/f_-(t)=\exp(1.4t-0.98)$ increases with $t$ (see proof of P3). Given that the  ROC curve is concave and $M_+=M_-=\infty$, P3 says the PR curve is nonincreasing, and by P2 the PR curve starts at (0,1) and ends at $(1,\pi_+)$. PR curves are shown for $\pi_+=1/2$ and $\pi_+=1/11$, to demonstrate the impact of imbalance; the $-$ class is ten times as likely as the $+$ class when $\pi_+=1/11$.
	
	\hspace*{2em} This ranking algorithm performs better than random (exceeds the chance curve) for both ROC and PR curves over all thresholds. The PR curve is much closer to the perfect-separation curve when $\pi_+=1/2$ than when $\pi_+=1/11$.
	
	\item[Case A*:] Scores follow a bi-lognormal model, meaning scores from  the $+$ class are distributed as lognormal with parameters $\mu_+=1.4$ and $\sigma^2_+=1$, while scores from the $-$ class are distributed as lognormal with parameters $\mu_-=0$ and $\sigma^2_-=1$. Because the same log transformation can be applied to scores from both the $+$ and $-$ classes to simultaneously convert the bi-lognormal model to a bi-normal model, property P7 tells us the ROC and PR  curves will match those displayed in Figure~\ref{fig:caseAB} for the bi-normal model.
	
	\item[Case B:] Scores from the $-$ class are distributed as normal with mean $\mu_-=0$ and variance $\sigma^2_-=1$. Scores from the $+$ class are distributed as $8-W$, where $W$ follows a lognormal distribution with parameters $\mu_+=1.4$ and $\sigma^2_+=1$. Letting $\Phi(t)$ and $\Phi^{-1}(t)$ denote the standard normal distribution and inverse distribution functions, the class-level distribution functions are $F_-(t)=\Phi(t)$ and $F_+(t)=\Phi([\mu_+-\log(8-t)]/\sigma_+)$, with $m_-=m_+=-\infty$, $M_+=8$, and $M_-=\infty$. By P1, the ROC curve starts at (0,0), is nondecreasing, and ends at (1,1). But the ROC curve is neither concave nor convex because $f_+(t)/f_-(t)=\exp(\left\{t^2-[1.4-\log(8-t)]^2\right\}/2)/(8-t)$ is not a monotone function of $t$. The PR curve is non-monotone, starting at (0,0) (by P2), increasing, then decreasing, then increasing again to end at $(1,\pi_+)$ (by P2).
	
	\hspace*{2em} This ranking algorithm is near-perfect for large thresholds (i.e., $t>0.7$, which results in $x_{\mathrm{ROC}}<0.2$, $y_{\mathrm{ROC}}$ and $x_{\mathrm{PR}}$ as large as 0.7, and $y_{\mathrm{PR}}>0.8$ when $\pi_+=0.5$), but is worse than random for very small thresholds (i.e., $t<-1$). One could argue, however, that the region where it performs poorly ($x_{\mathrm{ROC}}>0.75$) is of limited practical relevance because one would rarely consider using a threshold that results in a 75\% false-positive rate. Comparing performance of ranking algorithms from Cases A and B, ranking algorithm B is clearly better in regions of practical relevance, especially considering the PR curve when $\pi_+=1/11$. The choice of thresholds is clearly a critical component to comparing and choosing among ranking algorithms.

\begin{figure}
	\includegraphics[width=\textwidth]{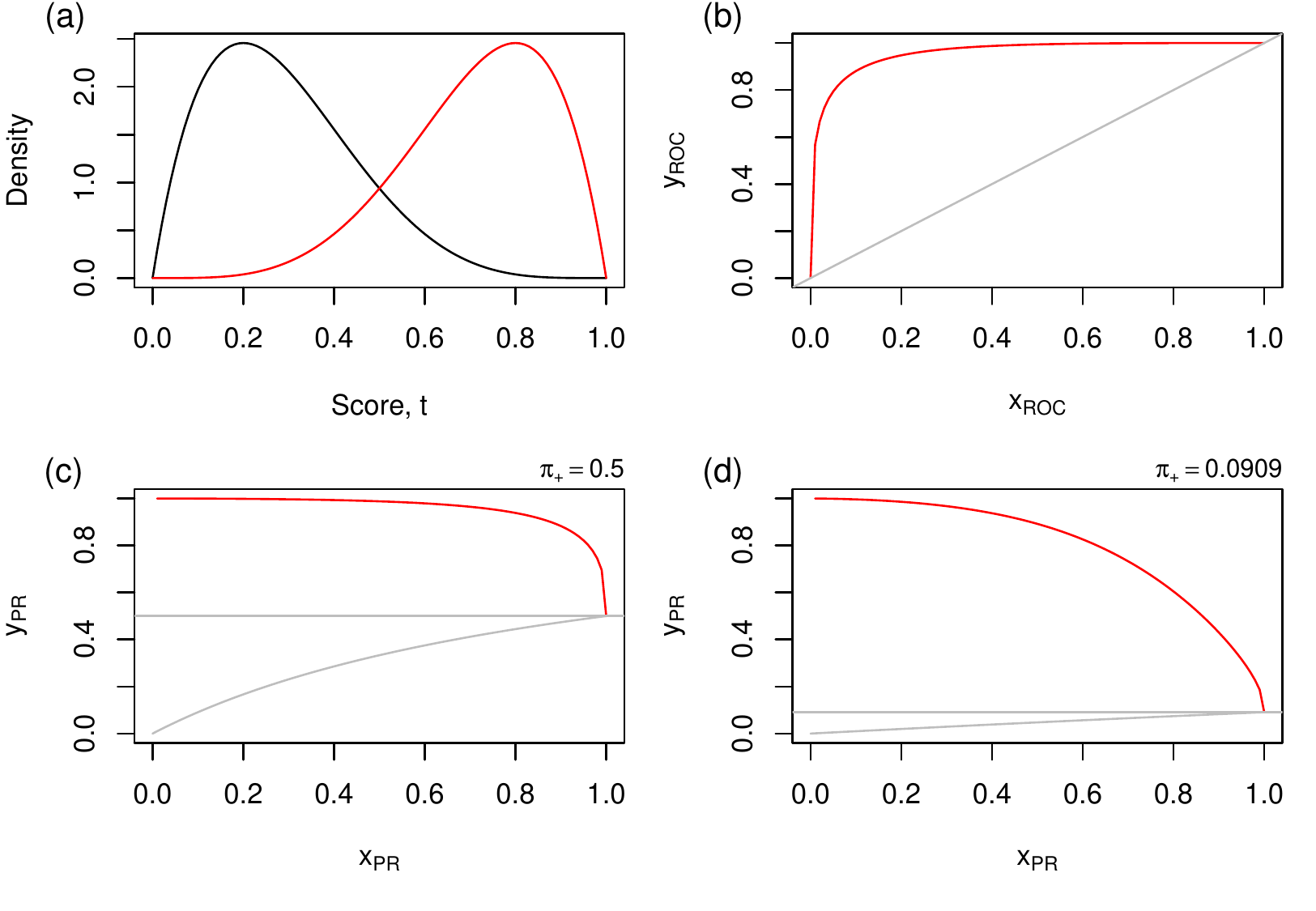}
	\caption{\label{fig:caseC} Densities, ROC curve, and PR curves for Case C. (a) Density functions for: the beta(2,5) distribution as the black curve, corresponding to the $-$ class; and for the beta(5,2) distribution as the red curve, corresponding to the $+$ class. (b) ROC curve, where the gray diagonal line is the chance curve. (c) PR curve assuming balanced classes so that $\pi_+=1/2$. The gray horizontal line is the chance curve, while the other gray curve is the achievable lower bound curve. (d) Same as (c), except assuming the $-$ class is ten times as likely as the $+$ class so that $\pi_+=1/11$. }
\end{figure}
	
	\item[Case C:] Scores follow a bi-beta model with scores for the $-$ class distributed as beta with parameters 2 and 5, and the $+$ class distributed as beta with parameters 5 and 2. Both ROC and PR curves are well behaved. The ROC curve starts at (0,0), increases to (1,1), and is concave because $f_+(t)/f_-(t)=[t/(1-t)]^3$ increases with $t$, where $m_-=m_+=0$ and $M_-=M_+=1$. As a result, the PR curve is decreasing from its starting point of (0,1) (by P3) to end at $(1,\pi_+)$ (by P2). The bi-beta model is particularly well-suited to scenarios where scores are bounded, and the beta distribution offers a variety of shapes (e.g., levels of skewness).

\begin{figure}
	\includegraphics[width=\textwidth]{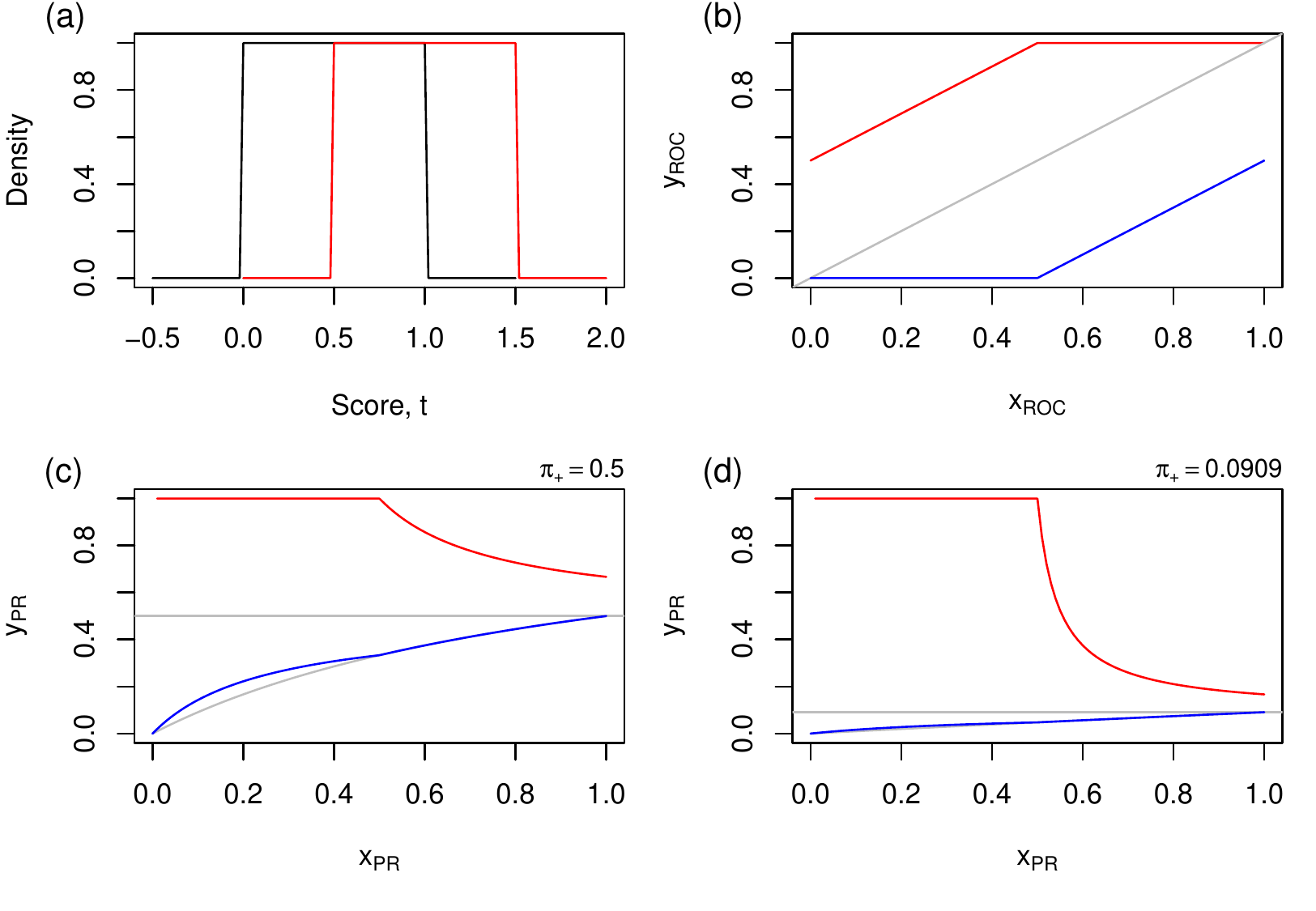}
	\caption{\label{fig:caseDE} Densities, ROC curves, and PR curves for Cases D and E. (a) Density functions for the uniform [0,1] distribution as the black curve and the uniform [0.5,1.5] as the red curve; vertical reference lines delineate the possible values for each density. For case D, the black curve corresponds to the $-$ class and the red curve to the $+$ class. Case E is the reverse of Case D. (b) ROC curves: Case D is the red curve, Case E is the blue curve. The gray diagonal line is the chance curve. (c) PR curves assuming balanced classes so that $\pi_+=1/2$: Case D is the red curve, Case E is the blue curve. The gray horizontal line is the chance curve, while the other gray curve is the achievable lower bound curve. (d) Same as (c), except assuming the $-$ class is ten times as likely as the $+$ class so that $\pi_+=1/11$. }
\end{figure}

	\item[Case D:] Scores have non-subset ranges, with scores for the $-$ class distributed as uniform on $[0,1]$ and scores for the $+$ class distributed as uniform on $[0.5,1.5]$. By P1, the ROC curves starts at $(0,0.5)$, which might at first seem strange but is a consequence of the overlapping but non-subset relationship between the possible scores for different classes. More specifically, the ROC curve does not start at (0,0) because $M_-=1<1.5=M_+$. Also by P1, the ROC curve is nondecreasing and ends at (1,1). The ROC curve is concave because $f_+(t)/f_-(t)=0$ for $0\leq t<0.5$, 1 for $0.5\leq t \leq 1$, $\infty$ for $1<t\leq 1.5$ is nondecreasing in $t$. Consequently, the PR curve is nonincreasing by P3 and by P2 it starts at (0,1) and ends at $(1,\frac{2\pi_+}{\pi_++1})$. The PR curve does not end at $(1,\pi_+)$ as with the previous cases because $m_-=0<0.5=m_+$, resulting in $F_-(m_+)=0.5$. The endpoint is (1,2/3) for balanced classes, and (1,1/6) when the $-$ class is ten times more likely than the $+$ class.
	
	\item[Case E:] This is Case D, except the scores are reversed for the classes. That is, scores for the $-$ class are distributed as uniform on $[0.5,1.5]$, and scores for the $+$ class are distributed as uniform on [0,1]. Operationally, large scores should suggest the $+$ class, so this ranking algorithm is expected to perform rather poorly. The ROC curve is still nondecreasing, but it is now below the chance curve, indicating a completely ineffective ranking algorithm (which is as expected). The ROC curve starts at (0,0) and ends at (1,0.5) by P1; it does not end at (1,1) because $m_-=0.5>0=m_+$. The ROC curve is convex because $f_+(t)/f_-(t)=\infty$ for $0\leq t<0.5$, 1 for $0.5\leq t \leq 1$, 0 for $1<t\leq 1.5$ is nonincreasing in $t$. By P3, the PR curve is nondecreasing because the ROC curve is convex, $M_+=1<1.5=M_-$, and $\lim_{t\uparrow M_+} \left\{ \frac{f_-(t)}{f_+(t)}\left[1-F_+(t)\right] \right\}=0$. By P2, the PR curve starts at (0,0) and ends at (1,$\pi_+$).

	\begin{figure}
		\includegraphics[width=\textwidth]{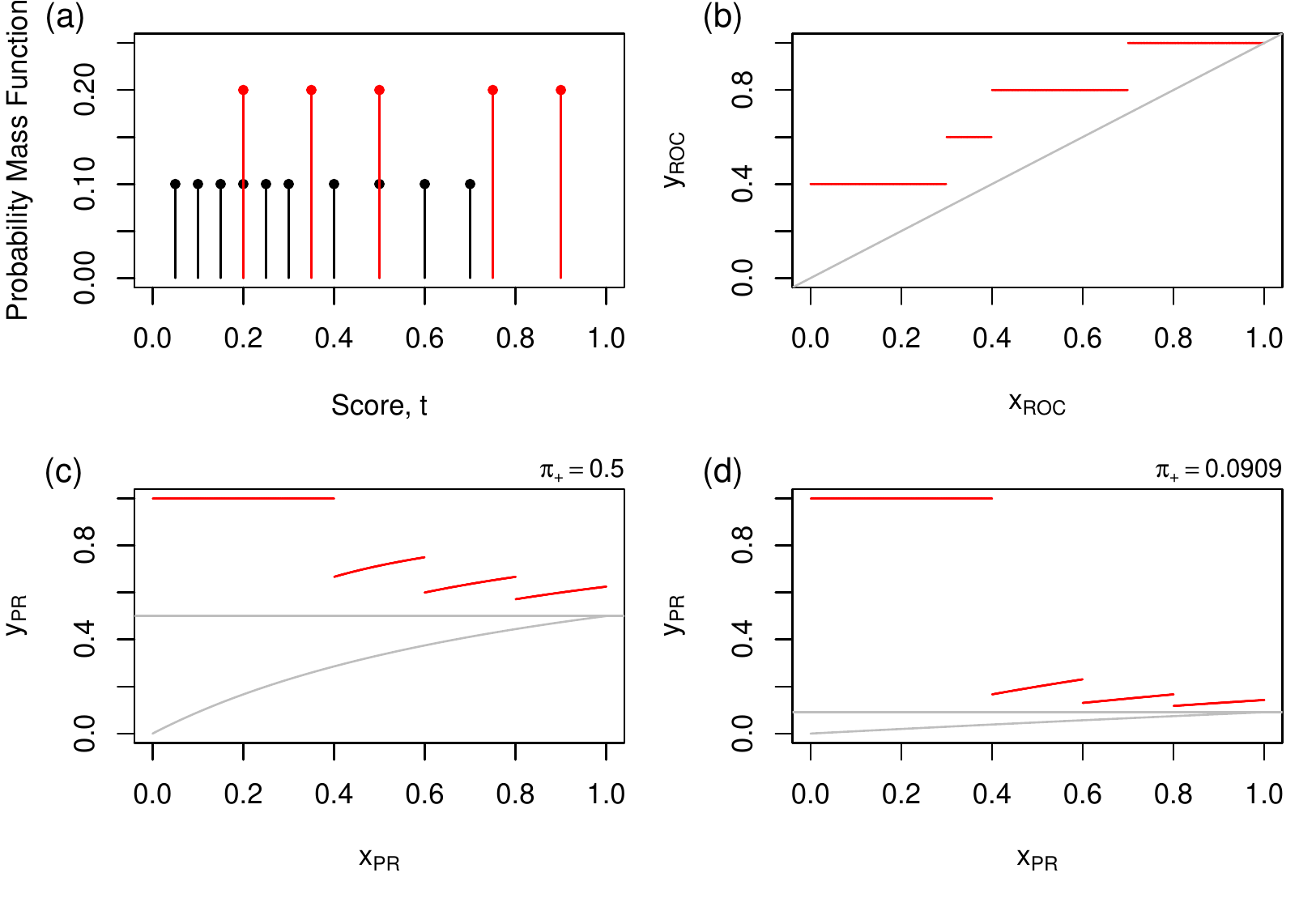}
		\caption{\label{fig:caseF} Probability mass functions, ROC curve, and PR curves for Case F. (a) Probability mass functions: the discrete uniform on 0.05, 0.1,.15, 0.2, 0.25, 0.3, 0.4, 0.5, 0.6, 0.7 as the black dots, corresponding to the $-$ class; and discrete uniform on  0.2, 0.35, 0.5, 0.75, 0.9 as the red dots, corresponding to the $+$ class. (b) ROC curve, where the gray diagonal line is the chance curve. (c) PR curve assuming balanced classes so that $\pi_+=1/2$. The gray horizontal line is the chance curve, while the other gray curve is the achievable lower bound curve. (d) Same as (c), except assuming the $-$ class is ten times as likely as the $+$ class so that $\pi_+=1/11$. }
	\end{figure}
	
	\item[Case F:] This case demonstrates ranking-algorithm scores where only a finite number of possible values are allowed, i.e., the scores are discrete random variables. Scores for the $-$ class follow a discrete uniform distribution with 10 possible score values of 0.05, 0.1, 0.15, 0.2, 0.25, 0.3, 0.4, 0.5, 0.6, and 0.7. Scores for the $+$ class follow a discrete uniform distribution with five possible score values of 0.2, 0.35, 0.5, 0.75, and 0.9. Because $m_-=0.05<0.2=m_+$ and $M_-=0.7<0.9=M_+$, P1 implies the ROC curve is nondecreasing, starting at $(0,0.4)$ and ending at $(1,1)$. Nothing can be said about monotonicity of the PR curve because the ROC curve is neither concave or convex. In fact, it can be seen that the PR curve is not monotone; it consists of  continuous nondecreasing pieces separated by jumps. By P2, the PR curve starts at (0,1) and ends at $(1,\frac{\pi_+}{\pi_++0.6})$. The PR curve endpoint is (1,0.625) for balanced classes, and (1,0.143) when the $-$ class is ten times more likely than the $+$ class. 
	
	\hspace*{2em} This case provides a preliminary view of what to expect from empirical estimation of the PR curve. Observed scores, even from continuous distributions, will be countable sets that may be viewed as arising from discrete distributions. Section~\ref{sec:emp.def.asymp} further addresses this connection.

\end{description}

\section{Empirical PR Curves}
\label{sec:empirical}
In this section, we define empirical PR curves as obtained from observed scores without additional distributional assumptions. Stochastic convergence properties are presented and investigated using small-sample simulation studies.

\subsection{Definitions and Asymptotics}
\label{sec:emp.def.asymp}

Suppose we observe independent random samples of scores $\{S^+_i;\ i=1,\ldots,n_+\}$ from the $+$ class and  $\{S^-_i;\ i=1,\ldots,n_-\}$ from the $-$ class, where $n=n_++n_-$ is the total number of items or instances. Let $\widehat{F}_+(\cdot)$ and $\widehat{F}_-(\cdot)$ denote the class-level empirical distribution functions, i.e., $\widehat{F}_+(t)=\displaystyle\frac{1}{n_+}\sum_{i=1}^{n_+}I(S^+_i\leq t)$ and $\widehat{F}_-(t)=\displaystyle\frac{1}{n_-}\sum_{i=1}^{n_+}I(S^-_i\leq t)$, where $I(\cdot)$ is the indicator function. Ignoring class membership and combining all scores as $\{S_i;\ i=1,\ldots,n\}$, also define $\widehat{F}(t)=\displaystyle\frac{1}{n}\sum_{i=1}^{n}I(S_i\leq t)$.

A popular and computationally efficient empirical estimator of the PR curve results from estimating the pair {\it (recall\;,\;precision)} for different thresholds as described in~(\ref{eq:pr-from-roc}). A natural selection of threshold values is the set $T$ of distinct scores observed (collectively over both classes). In other words, an estimated PR curve is
\begin{eqnarray*} 
	\widehat{PR}^* &=& \left\{ \left(\mbox{proportion of $S^+$ values}> t, \frac{\mbox{number of $S^+$ values}> t}{\mbox{number of $S$ values}> t} \right);\  t\in T \right\}\\
	&=&  \left\{ \left(1-\widehat{F}_+(t), \frac{n_+[1-\widehat{F}_+(t)]}{n[1-\widehat{F}(t)]} \right);\  t\in T \right\}.
\end{eqnarray*}
To avoid division by zero while not using if-then-else statements, sometimes the slightly modified definition
\[ \widehat{PR}^0 = \left\{ \left(\mbox{proportion of $S^+$ values}\geq t, \frac{\mbox{number of $S^+$ values}\geq t}{\mbox{number of $S$ values}\geq t} \right);\  t\in T \right\} \]
is used. Neither $\widehat{PR}^*$ nor $\widehat{PR}^0$ present as functional definitions for estimated precision as a function of recall. More importantly, they can result in multiple distinct values of estimated precision for a single value of recall. This creates confusion when evaluating metrics such as area under the PR curve. \citet{DavisGoadrich2006} and \citet{BoydEngPage2013}  extensively investigate this issue. They also give guidance on the proper {\it interpolation} that should be used to fill in the gaps created in the estimated PR curve based on estimators $\widehat{PR}^*$ and $\widehat{PR}^0$. Clearly, a functional definition would avoid the need for interpolation.

This author sides with \citet{ClemenconVayatis2009} in recommending the estimator that naturally comes from the functionally defined PR curve as presented in~(\ref{eq:prcdfs}), but applied using $\widehat{F}_+$ and $\widehat{F}_-$ in place of $F_+$ and $F_-$. Specifically, consider the empirical estimator of the PR curve
\begin{eqnarray*} 
	\widehat{PR} 
	&=&  \left\{ \left(x, \frac{(n_+/n)}{(n_+/n) +\displaystyle\frac{n_-/n}{x}\left[1-\widehat{F}_-\left(\widehat{F}_+^{-1}(1-x)\right)\right]} \right), 0< x\leq 1 \right\} \\
	&=&  \left\{ \left(x, \frac{1}{1 +\displaystyle\frac{n_-/n_+}{x}\left[1-\widehat{F}_-\left(\widehat{F}_+^{-1}(1-x)\right)\right]} \right), 0< x\leq 1 \right\}.
\end{eqnarray*}
This estimator is free of the disadvantages given above for $\widehat{PR}^*$ and $\widehat{PR}^0$. While it requires computation of the inverse empirical distribution function $\widehat{F}_+^{-1}(\cdot)$, also known as the empirical quantile function, the additional computations are well worth the benefits. 

Some properties of $\widehat{PR}$ are clearly demonstrated in Figure~\ref{fig:caseF} showing Case~F based on discrete distributions for scores. Sampling, even from continuous score populations, will result in empirical distribution functions that are step functions as in Case~F. The resulting $\widehat{PR}$ curve consists of at most $n_+$ disjoint segments, each associated with distinct values of observed $S^+$. Suppose the points of discontinuity occur at $x_1,\ x_2,\ \ldots,\ x_m$. Then the $\widehat{PR}$ segments are defined on $x\in(0,x_1),\ x\in[x_1,x_2),\ \ldots$, and $x\in[x_m,1)$. $\widehat{PR}(\cdot)$ is either continuous or continuous from the right at $x$. Estimated precision is an increasing function on each segment. Furthermore, at points of discontinuity, the limit from the left of the estimated precision curve is larger than the limit from the right, i.e., $\lim_{x\uparrow x_j}\widehat{PR}(x)>\widehat{PR}(x_j)$. 

An extensive body of literature focuses on asymptotic properties of the empirical ROC curve $1-\widehat{F}_+\left(\widehat{F}_-^{-1}(1-x)\right)$; see, for example, \citet{Csorgo1983}, \citet{HsiehTurnbull1996}, \citet{Pepe2003}, and \citet{Bertailetal2009}. Building on this body of literature to take advantage of the similarity between the empirical ROC and $1-\widehat{F}_-\left(\widehat{F}_+^{-1}(1-x)\right)$ as appearing in $\widehat{PR}$, then applying a multivariate Taylor approximation to the function that converts $1-\widehat{F}_-\left(\widehat{F}_+^{-1}(1-x)\right)$ to obtain $\widehat{PR}$, \citet{ClemenconVayatis2009} obtain the following strong approximation result for $\widehat{PR}$.

\begin{theorem}[Strong approximation]
Suppose $F_+$ and $F_-$ have densities $f_+$ and $f_-$, and the following conditions hold:
\begin{enumerate}
	\item For some $\epsilon>0$, the slope of the function $x\mapsto 1-{F}_-\left({F}_+^{-1}(1-x)\right)$ is bounded on $[\epsilon,1-\epsilon]$, i.e.,
	\begin{equation}
	\label{eq:slope}
	\sup_{x\in[\epsilon,1-\epsilon]} \frac{f_-\left({F}_+^{-1}(1-x)\right)}{f_+\left({F}_+^{-1}(1-x)\right)} <\infty.
	\end{equation}
	
	\item The density $f_+$ 
	\begin{enumerate}
		\item is differentiable,
		\item does not vanish for $x\in[\epsilon,1-\epsilon]$, i.e., 
		\begin{equation}\nonumber
		f_+\left({F}_+^{-1}(1-x)\right)>0 \quad \mbox{for all} \quad x\in[\epsilon,1-\epsilon],
		\end{equation}
		\item and has controlled tail behavior in that there exists $\gamma>0$ such that
		\begin{equation}
		\label{eq:tail}
		\sup_{x\in[\epsilon,1-\epsilon]} x(1-x) \left|\frac{\mbox{d}}{\mbox{d}x}\log\left(f_+\left({F}_+^{-1}(1-x)\right)\right)\right| \leq \gamma<\infty.
		\end{equation}
	\end{enumerate}  
\end{enumerate}
Then, we almost surely have, as $n\rightarrow\infty$:
\begin{enumerate}
	\item[A.] The empirical PR curve is strongly consistent, uniformly over $[\epsilon,1-\epsilon]$, i.e.,
	\[ \sup_{x\in[\epsilon,1-\epsilon]} \left|\widehat{PR}(x)-PR(x) \right| \rightarrow 0 . \]
	\item[B.] There exist two independent sequences of Brownian bridges $\{B_1^{(n)}(x)\}_{x\in(0,1)}$ and \\ $\{B_2^{(n)}(x)\}_{x\in(0,1)}$, and a Gaussian random variable $W$ independent from the Brownian bridges, such that uniformly over $[\epsilon,1-\epsilon]$:
	\begin{eqnarray}
	\nonumber
	\sqrt{n}\left(\widehat{PR}(x)-PR(x) \right) &=& Z^{(n)}(x) + o\left(\frac{(\log\log n)^{\rho_1(\gamma)}(\log n)^{\rho_2(\gamma)}}{\sqrt{n}}\right), \\ \label{eq:asymp}
	Z^{(n)}(x) &=& \displaystyle \frac{PR(x)^2}{x} \left(\frac{\sqrt{1-\pi_+}}{\pi_+^{3/2}}\right) \left[1-{F}_-\left({F}_+^{-1}(1-x)\right)\right] W  \\\nonumber
	&&\displaystyle +\  \frac{PR(x)^2}{x} \left(\frac{1-\pi_+}{\pi_+^{3/2}}\right) \left(\frac{f_-\left({F}_+^{-1}(1-x)\right)}{f_+\left({F}_+^{-1}(1-x)\right)}\right) B_1^{(n)}(x)  \\\nonumber
	&&\displaystyle +\  \frac{PR(x)^2}{x} \left(\frac{\sqrt{1-\pi_+}}{\pi_+}\right)  B_2^{(n)}\left(1-{F}_-\left({F}_+^{-1}(1-x)\right)\right), \\\nonumber
	&& \begin{cases} 
	\rho_1(\gamma)=0, \quad \rho_2(\gamma)=1, & \mbox{if } \gamma<1\\
	\rho_1(\gamma)=0, \quad \rho_2(\gamma)=2, & \mbox{if } \gamma=1\\
	\rho_1(\gamma)=\gamma, \quad \rho_2(\gamma)=\gamma-1+\epsilon,\ \epsilon>0, & \mbox{if } \gamma>1
	\end{cases} .
	\end{eqnarray}
\end{enumerate}
Moreover, pointwise limits are obtained for a fixed $x$ in $[\epsilon,1-\epsilon]$ as 
	\begin{equation*}
	\sqrt{n}\left(\widehat{PR}(x)-PR(x) \right) \stackrel{\mbox{d}}{\longrightarrow}{\cal{N}}(0,\sigma^2(x)), 	\end{equation*}
where
	\begin{eqnarray}\label{eq:avar}
	\lefteqn{\sigma^2(x) = \displaystyle \frac{PR(x)^4}{x^2}\cdot skew(1+skew)\cdot} \\\nonumber
		&&\left\{ \alpha^2(1+skew)  + \left[\frac{f_-\left({F}_+^{-1}(1-x)\right)}{f_+\left({F}_+^{-1}(1-x)\right)}\right]^2 x(1-x)\cdot skew  + \alpha(1-\alpha) \right\},	
	\end{eqnarray}
	and $skew=\frac{1-\pi_+}{\pi_+}$ and $\alpha=1-{F}_-\left({F}_+^{-1}(1-x)\right)$.
\end{theorem}
\noindent 
Note that a typo has been corrected in (\ref{eq:asymp}), namely $\sqrt{\pi_+}$ replaced a $\pi_+$ in the first term of $Z^{(n)}(x)$. Typos have also been corrected in (\ref{eq:avar}): the last two terms of $\sigma^2(x)$ were missing multipliers $PR(x)^4/x^2$, and inverse distribution functions were needed in two places. For defining tail behavior, $x(1-x)$ has been included in~(\ref{eq:tail}), in the spirit of \citet{Parzen1979}.

The variance decomposition presented in~(\ref{eq:avar}) has interesting properties. Variance clearly increases as $skew$ (i.e., imbalance) increases. The slope of the function $x\mapsto 1-{F}_-\left({F}_+^{-1}(1-x)\right)$ as given in~(\ref{eq:slope}) is important, with variance being a quadratic function of this slope; variance can quickly increase for large slopes. On the other hand, the slope being zero causes the second term of the variance to vanish. The first and third term of the variance vanish when ${F}_-\left({F}_+^{-1}(1-x)\right)=1$; this happens when scores have different ranges across classes, with the $+$ class having larger values. The third term of the variance also vanishes when ${F}_-\left({F}_+^{-1}(1-x)\right)=0$; this happens when scores have different ranges across classes, with the $+$ class having smaller values.

The normal approximations suggested by (\ref{eq:asymp}) work very well for some situations, even for relatively small $n$ and large $skew$. On the other hand, they are completely inappropriate in other situations. These comments are further discussed in the following subsection.

\subsection{Small-sample Properties}
\begin{sidewaysfigure}
	\includegraphics[width=\textwidth,trim={0 0 0 0.65cm},clip]{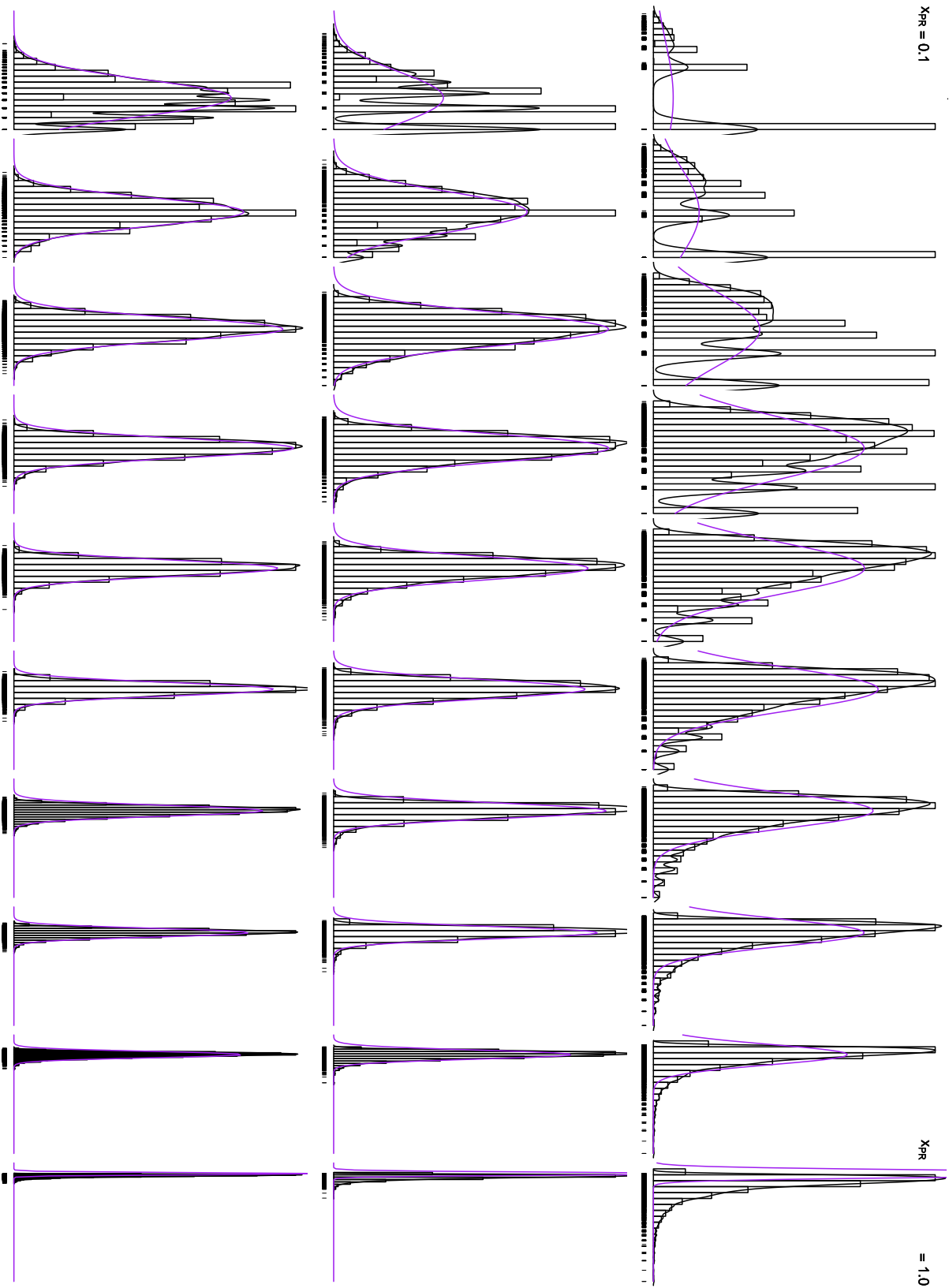}
	\caption{Histograms of observed values of $\widehat{PR}(x)$. Data is generated according to Case~A, with $skew=10$. Rows represent different sample sizes: $n=100$ on top, $n=500$ in the middle, $n=1000$ on the bottom. Columns represent values of $x$: $x=0.1$ (leftmost column), $0.2,\ldots,0.9,1$ (rightmost column). Each histogram is based on 5000 simulation replicates, and includes a kernel density estimate as the black curve. ``Rugs'' below each histogram give extra information on observed values, where the horizontal axis extends from zero to one. Approximating normal densities as obtained from~(\ref{eq:avar}) are also shown as the purple curve.}
	\label{fig:caseAskew10hist}  
\end{sidewaysfigure}

\begin{sidewaysfigure}
	\includegraphics[width=\textwidth,trim={0 0 0 0.65cm},clip]{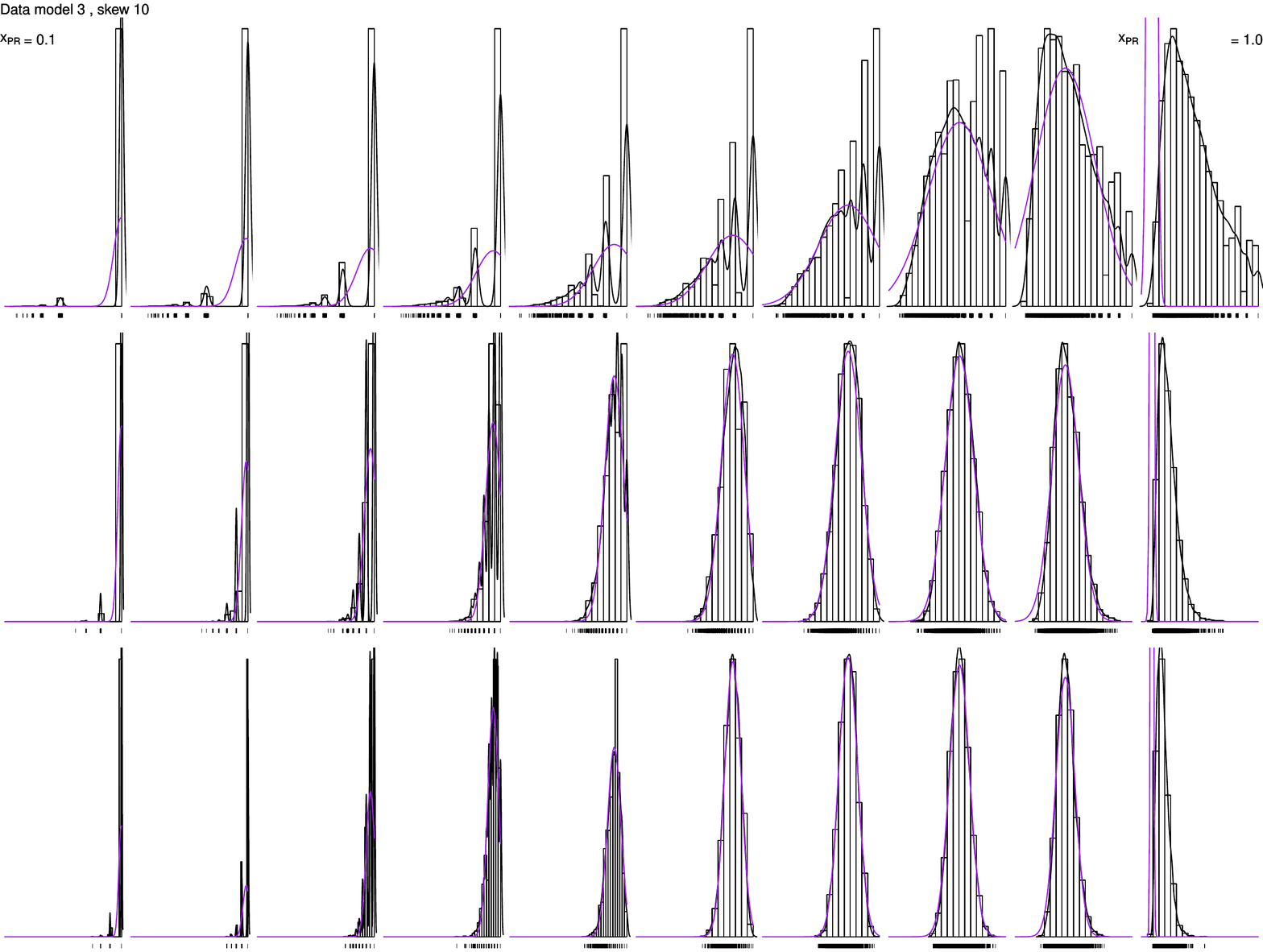}
	\caption{Histograms of observed values of $\widehat{PR}(x)$. Data is generated according to Case~C, with $skew=10$. Rows represent different sample sizes: $n=100$ on top, $n=500$ in the middle, $n=1000$ on the bottom. Columns represent values of $x$: $x=0.1$ (leftmost column), $0.2,\ldots,0.9,1$ (rightmost column). Each histogram is based on 5000 simulation replicates, and includes a kernel density estimate as the black curve. ``Rugs'' below each histogram give extra information on observed values, where the horizontal axis extends from zero to one. Approximating normal densities as obtained from~(\ref{eq:avar}) are also shown as the purple curve.}
	\label{fig:caseCskew10hist}  
\end{sidewaysfigure}

\begin{sidewaysfigure}
	\includegraphics[width=\textwidth,trim={0 0 0 0.65cm},clip]{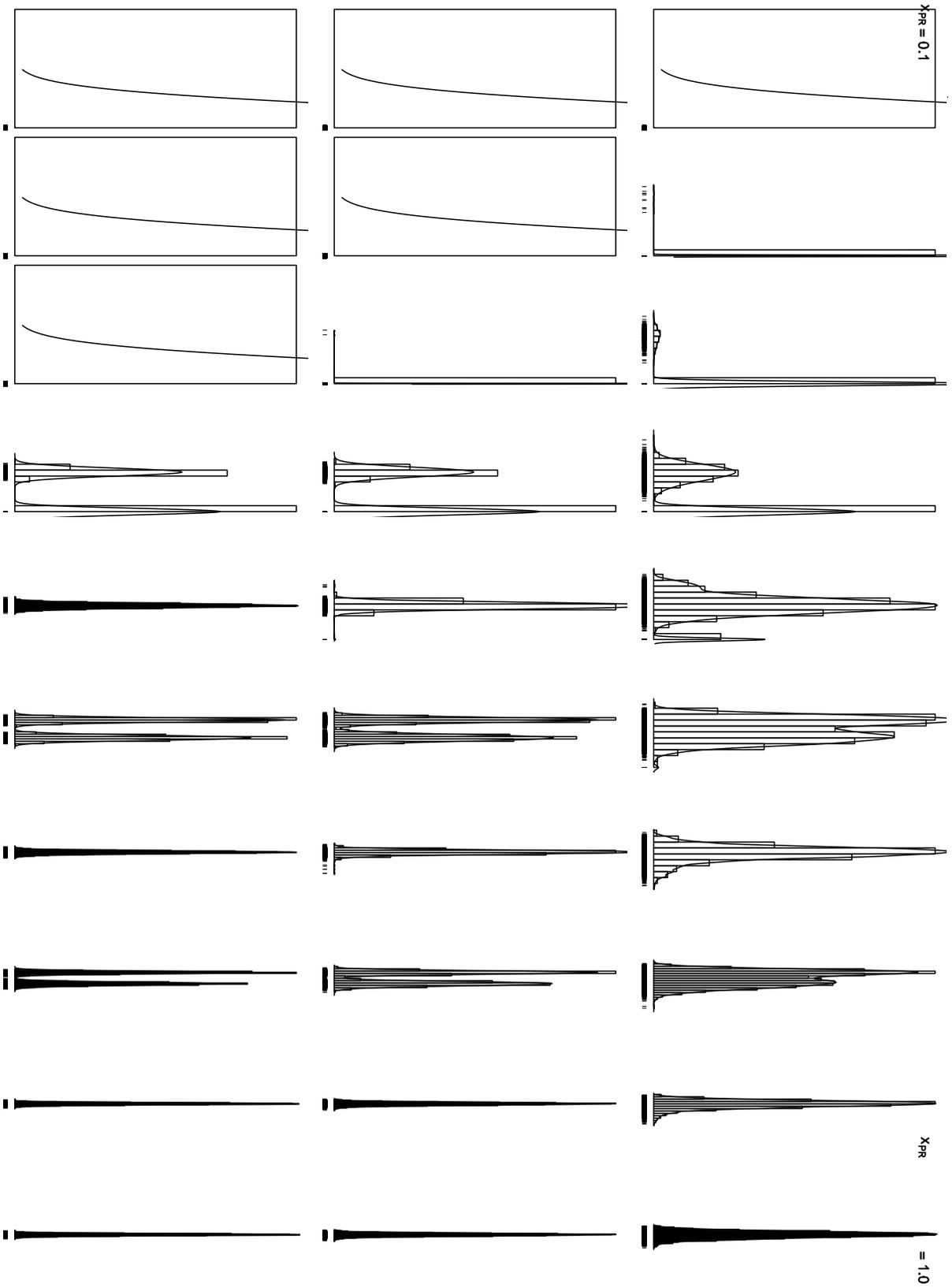}
	\caption{Histograms of observed values of $\widehat{PR}(x)$. Data is generated according to Case~F, with $skew=1$. Rows represent different sample sizes: $n=100$ on top, $n=500$ in the middle, $n=1000$ on the bottom. Columns represent values of $x$: $x=0.1$ (leftmost column), $0.2,\ldots,0.9,1$ (rightmost column). Each histogram is based on 5000 simulation replicates, and includes a kernel density estimate as the black curve. ``Rugs'' below each histogram give extra information on observed values, where the horizontal axis extends from zero to one. Note that when all observed values are nearly one, as in histograms for small $x$, this is clearly demonstrated by the rugs but the histogram appears as a single bin and the kernel density estimate is not good.}
	\label{fig:caseFskew1hist}  
\end{sidewaysfigure}

\begin{sidewaysfigure}
	\includegraphics[width=\textwidth,trim={0 0 0 0.1cm},clip]{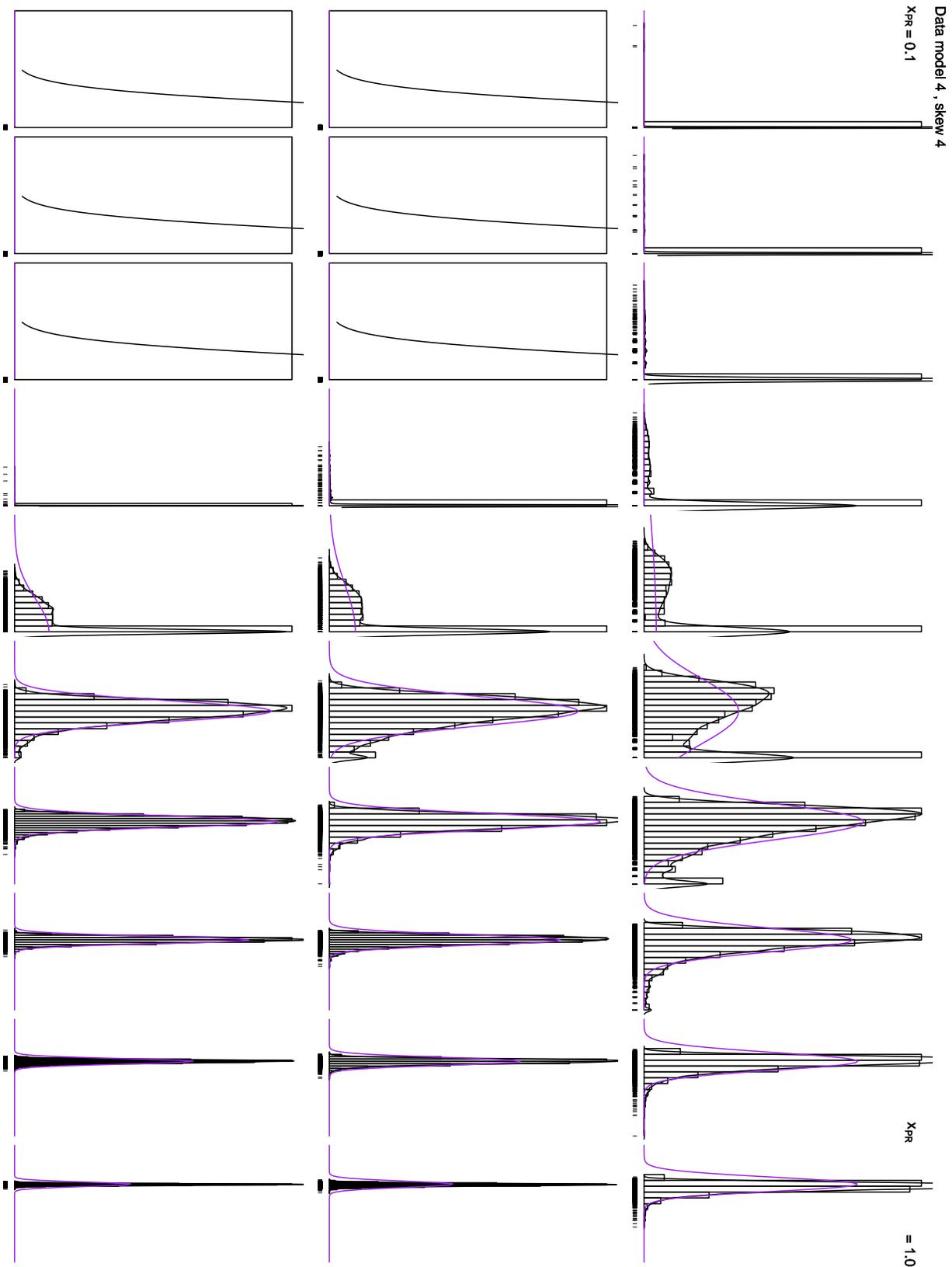}
	\caption{Histograms of observed values of $\widehat{PR}(x)$. Data is generated according to Case~D, with $skew=4$. Rows represent different sample sizes: $n=100$ on top, $n=500$ in the middle, $n=1000$ on the bottom. Columns represent values of $x$: $x=0.1$ (leftmost column), $0.2,\ldots,0.9,1$ (rightmost column). Each histogram is based on 5000 simulation replicates, and includes a kernel density estimate as the black curve. ``Rugs'' below each histogram give extra information on observed values, where the horizontal axis extends from zero to one. Note that when all observed values are nearly one, as in histograms for small $x$, this is clearly demonstrated by the rugs but the histogram appears as a single bin and the kernel density estimate is not good. Approximating normal densities as obtained from~(\ref{eq:avar}) are also shown as the purple curve.}
	\label{fig:caseDskew4hist}  
\end{sidewaysfigure}

\begin{sidewaysfigure}
	\includegraphics[width=\textwidth,trim={0 0 0 0.1cm},clip]{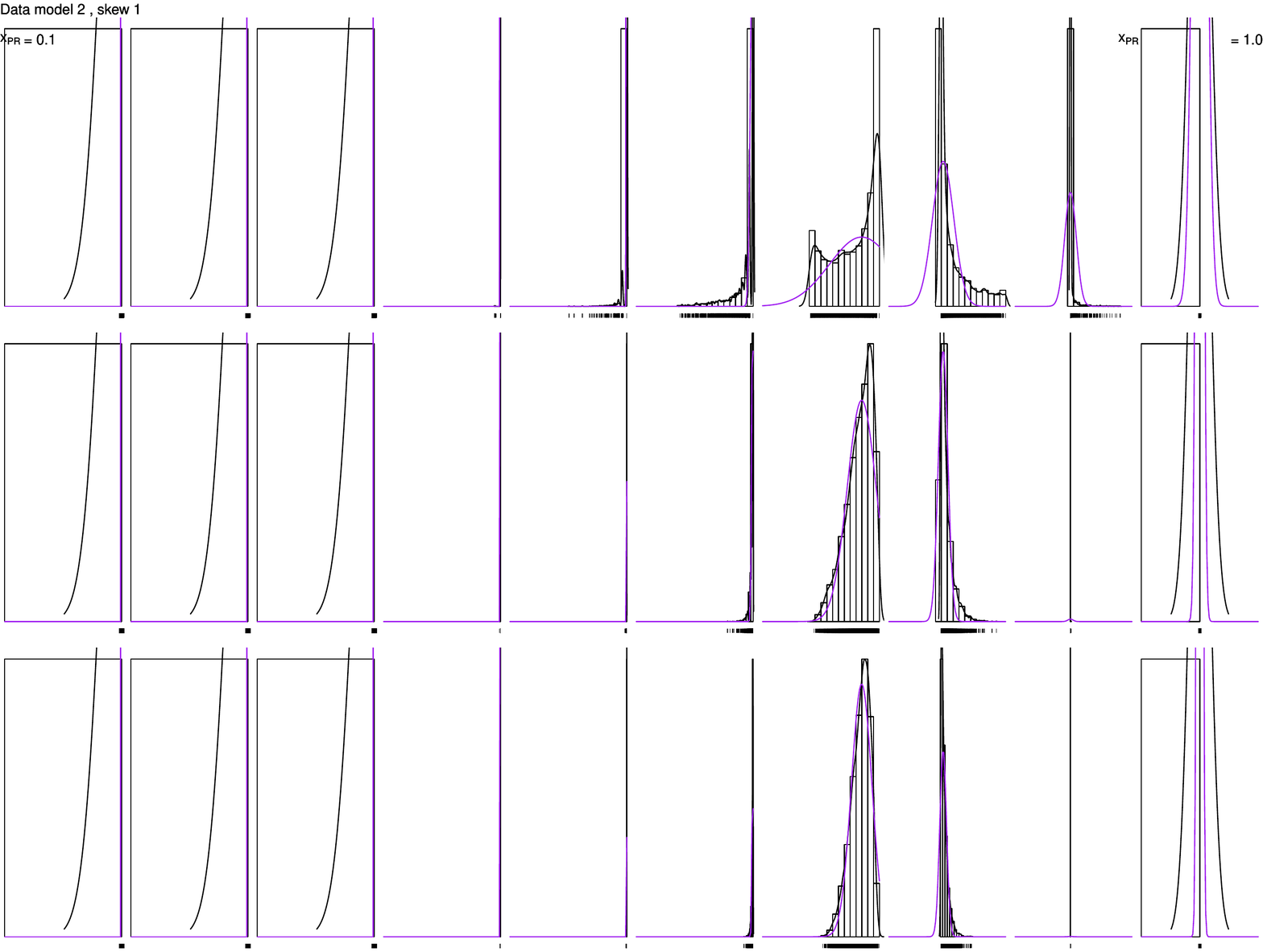}
	\caption{Histograms of observed values of $\widehat{PR}(x)$. Data is generated according to Case~B, with $skew=1$. Rows represent different sample sizes: $n=100$ on top, $n=500$ in the middle, $n=1000$ on the bottom. Columns represent values of $x$: $x=0.1$ (leftmost column), $0.2,\ldots,0.9,1$ (rightmost column). Each histogram is based on 5000 simulation replicates, and includes a kernel density estimate as the black curve. ``Rugs'' below each histogram give extra information on observed values, where the horizontal axis extends from zero to one. Note that when all observed values are nearly one, as in histograms for small $x$, this is clearly demonstrated by the rugs but the histogram appears as a single bin and the kernel density estimate is not good. Approximating normal densities as obtained from~(\ref{eq:avar}) are also shown as the purple curve.}
	\label{fig:caseBskew1hist}  
\end{sidewaysfigure}

To study the small-sample behavior of $\widehat{PR}(x)$, samples of sizes $n=100,\ 500,\ 1000$ using $skew=10,\ 4,\ 1$ (corresponding to $\pi_+=1/11, 0.2, 0.5$) were generated from Cases A, B, C, D, and F. Some of the resulting histograms, based on 5000 simulation replicates, of $\widehat{PR}(x)$ are shown in Figures~\ref{fig:caseAskew10hist}--\ref{fig:caseBskew1hist}, at $x=0.1,0.2,\ldots,1$. For a particular data-generating mechanism and a value for $skew$, these figures show histograms of $\widehat{PR}(0.1)$, $\widehat{PR}(0.2)$, \ldots, $\widehat{PR}(1.0)$ for $n=100,\ 500,\ 1000$. A kernel density estimate is shown along with each histogram, and in some figures  approximating normal densities as obtained from (\ref{eq:avar}) are also shown. In situations where the normal approximation is valid, we expect the kernel density to coincide with the normal density, with improved performance for increasing $n$, where the top row of histograms correspond to $n=100$ and the bottom row to $n=1000$.

Consider Case A where scores have the same range for both classes (meaning $F_+$ and $F_-$ have equal support), and class-specific densities have exponential-type tail behavior with $\gamma=1$ in~(\ref{eq:tail}). The slope in~(\ref{eq:slope}) gets very large as $x$ approaches 1, but is less than 5 when $x\leq 0.9$. These are near-ideal conditions for the normal approximation to hold. Figure~\ref{fig:caseAskew10hist} demonstrates that even with a large $skew=10$ and relatively small $n=500$, resulting in a small $n_+=45$ on average, the normal approximation from (\ref{eq:avar}) very nearly matches the kernel density when $0.2\leq x\leq 0.9$. The normal approximation does not perform well near the extremes, namely $x$ close to zero or one; this is no surprise as the approximation in (\ref{eq:asymp}) is valid for $x\in[\epsilon,1-\epsilon]$ for $\epsilon>0$.

For other situations, much larger $n$ is needed to deal with the same value of $skew=10$. Consider Case C where scores again have the same range for both classes. A major difference from Case A is that the scores of Case C have finite range, corresponding to $\gamma<1$ which by itself would yield faster convergence. However, the slope from~(\ref{eq:slope}) can be very large, especially for $x>0.9$. This will serve to destabilize the second term of $Z^{(n)}(x)$ in~(\ref{eq:asymp}). The second term of $Z^{(n)}(x)$ in~(\ref{eq:asymp}) also suggests that decreasing the $skew$ may offset the effect of large slope. In fact, all three terms in $Z^{(n)}(x)$ are expected to become more stable as $skew$ approaches one. Additionally, $1-{F}_-\left({F}_+^{-1}(1-x)\right)\leq 0.0061$ when $x\leq0.5$, essentially dropping the first and third terms from $Z^{(n)}(x)$. All of this results in slower convergence, as demonstrated in Figure~\ref{fig:caseCskew10hist}. 

The shortcomings and limitations of Theorem~1 are quite enlightening. First, the results do not apply to scores observed from discrete populations. This, of course, is obvious because the theorem calls for densities that do not exist. However, other limitations exist. Figure~\ref{fig:caseFskew1hist} corresponding to Case~F demonstrates that $\widehat{PR}(x)$ may be inconsistent. For the ``best'' scenario of $skew=1$ and $n=1000$, the distributions of $\widehat{PR}(0.4)$, $\widehat{PR}(0.6)$, and $\widehat{PR}(0.8)$ are all bimodal. As may be seen in Figure~\ref{fig:caseF}, $PR(x)$ is discontinuous at exactly the same values of $x=0.4,\ 0.6,\ 0.8$. The root cause of inconsistency is that $\widehat{F}^{-1}_+(1-x)$ is only consistent for $F^{-1}_+(1-x)$ provided $F^{-1}_+(1-x)$ is continous at $1-x$~\citep[p.5]{Csorgo1983}. Scores from discrete distributions violate this condition.

Theorem~1 may be thoughtfully applied even when scores do not have the same range for both classes. Consider Case~D. When $x<0.5$, then $F^{-1}_+(1-x)>1$ and consequently, ${F}_-\left({F}_+^{-1}(1-x)\right)=1$ and ${f}_-\left({F}_+^{-1}(1-x)\right)=0$. Hence, (\ref{eq:asymp}) yields $Z^{(n)}(x)=0$, so the limiting distribution is degenerate. Moreover, the difficulty of estimating the boundary $(x=0.5)$ where the class densities no longer overlap has the consequence that larger $n$ and smaller $skew$ will be needed for the normal approximation to be reasonable even when $x\geq 0.5$. See Figure~\ref{fig:caseDskew4hist} for Case~D with $skew=4$. Even when $n=1000$, the approximation is reasonable only for $x>0.5$.

The very poor performance of the normal approximation in Case B may be somewhat surprising. See Figure~\ref{fig:caseBskew1hist} for Case~B when $skew=1$; the approximation is far worse for other values of skew. The tail behavior of the lognormal distribution yields $\gamma=1.25$, resulting in the slowest convergence rate among all cases considered in this article. When $x\leq 0.5$, both $1-{F}_-\left({F}_+^{-1}(1-x)\right)$ and $\frac{f_-\left({F}_+^{-1}(1-x)\right)}{f_+\left({F}_+^{-1}(1-x)\right)}$ are essentially zero, causing all three terms in $Z^{(n)}(x)$ to practically drop out. When $x\geq 0.9$, $\frac{f_-\left({F}_+^{-1}(1-x)\right)}{f_+\left({F}_+^{-1}(1-x)\right)}$ is again neglible, causing the second term in $Z^{(n)}$ to practically drop out. Also when $x\geq 0.9$, $1-{F}_-\left({F}_+^{-1}(1-x)\right)$ is so close to one that the Brownian bridge $B_2^{(n)}(\alpha)$ is essentially degenerate, with practical consequence that $Z^{(n)}(x)$ basically has only the first term.

\section{Concluding Remarks} 


This paper contains a comprehensive exposition on properties of population PR curves. Some results have been previously presented, most notably by \citet{DavisGoadrich2006}, \citet{ClemenconVayatis2009}, and by \citet{Boydetal2012}. Other results are new or conditions have been relaxed. By looking at a variety of distributional settings, defined according to Cases A to F, new results have been discovered.

This paper also investigates properties of the functional empirical estimator $\widehat{PR}$ of the PR curve. It is quite alarming that $\widehat{PR}(x)$ is not consistent at points corresponding to positive probability from discrete distributions of scores in the $+$ class. For continuously-defined scores, strong approximation is useful but convergence rates can be heavily influenced by the distributional setting, the $skew$, and the point of interest $x$ on the PR curve.

While the population and empirical PR curves inherit many properties from their ROC counterparts, PR curves have several complexities not seen in ROC curves. A thorough understanding of these complexities will allow users to avoid misuse or misinterpretation.

\appendix
\section*{Appendix A}
Properties of ROC and PR curves as given in Section~\ref{sec:properties} are proved below.

\begin{itemize}
	\item[P1.] 
	\begin{proof} 
		By equation (\ref{eq:roccdfs}), $y_{\mathrm{ROC}}\left( x_{\mathrm{ROC}} \right)= 1- F_+\left( F_-^{-1}(1-x_{\mathrm{ROC}})\right)$ for $0<x_{\mathrm{ROC}}<1$. As stated above, distribution functions and their generalized inverses are nondecreasing functions. Hence, $F_-^{-1}(1-x_{\mathrm{ROC}})$ is nonincreasing in $x_{\mathrm{ROC}}$, so that $F_+\left( F_-^{-1}(1-x_{\mathrm{ROC}})\right)$ is also nonincreasing in $x_{\mathrm{ROC}}$, and finally $1-F_+\left( F_-^{-1}(1-x_{\mathrm{ROC}})\right)$ is nondecreasing in $x_{\mathrm{ROC}}$  for $0<x_{\mathrm{ROC}}<1$.
		
Noting that $\lim_{x\downarrow0}F_-^{-1}(x)=m_-$ and $\lim_{x\uparrow1}F_-^{-1}(x)=M_-$, we get $\lim_{x_{\mathrm{ROC}}\downarrow0}y_{\mathrm{ROC}}\left( x_{\mathrm{ROC}} \right)=1-F_+\left( \lim_{x_{\mathrm{ROC}}\downarrow0} F_-^{-1}(1-x_{\mathrm{ROC}})\right)=1-F_+(M_-)$. Similarly, $\lim_{x_{\mathrm{ROC}}\uparrow1}y_{\mathrm{ROC}}\left( x_{\mathrm{ROC}} \right)=1-F_+\left( \lim_{x_{\mathrm{ROC}}\uparrow1} F_-^{-1}(1-x_{\mathrm{ROC}})\right)=1-F_+(m_-)$.
	\end{proof}
	
	\item[P2.]  
	\begin{proof} To show possible non-monotonicity, consider continuous ranking-algorithm scores that admit densities.
		From equation (\ref{eq:prcdfs}),
	\begin{equation}
	y_{\mathrm{PR}}(x_{\mathrm{PR}}) = 
	\frac{\pi_+}{\pi_+ + \pi_- g(x_{\mathrm{PR}}) }, \qquad 0 < x_{\mathrm{PR}} < 1,
	\label{eq:p2a}
	\end{equation}
	where $g(x)=\left[ 1-F_-\left( F_+^{-1}(1-x) \right) \right]/x$. Monotonicity of the PR curve is determined by monotonicity of $g(x)$ for a fixed value of $\pi_+$. Because
	\begin{eqnarray}
	g'(x) &=& \frac{1}{x^2}\left\{ \frac{-f_-\left( F_+^{-1}(1-x)\right)}{-f_+\left( F_+^{-1}(1-x)\right)} \cdot x - \left[ 1- F_-\left(F_+^{-1}(1-x)\right) \right] \right\} \nonumber \\ 
	&=& \frac{1}{x^2}\left\{ \frac{f_-\left( t \right)}{f_+\left( t\right)} \cdot \left[1-F_+(t)\right] - \left[ 1- F_-\left(t\right) \right] \right\}, \quad t=F_+^{-1}(1-x) \label{eq:p2b}
	\end{eqnarray}
	can be positive or negative, the PR curve can decrease or increase as $x_{\mathrm{PR}}$ changes for a fixed $\pi_+$. Moreover, 
	\[ \lim_{x\uparrow1}g(x) = 1-F_-\left( \lim_{x\uparrow1} F_+^{-1}(1-x)\right)=1-F_-(m_+) \]
	yields (a). 	When $M_+ < M_-$,
	\[ \lim_{x\downarrow0}g(x) = \frac{\lim_{x\downarrow0} \left[ 1-F_-\left( F_+^{-1}(1-x) \right)\right]}{\lim_{x\downarrow0} \ \ x} = \frac{\left[ 1-F_-\left( M_+ \right)\right]}{\lim_{x\downarrow0} \ \ x} =\infty \]
	because $1-F_-\left( M_+ \right)>0$, and this yields (c).
	Because $1-F_-\left( M_+ \right)=0$ when $M_+\geq M_-$, using l'Hopital's Rule we get
	\[ \lim_{x\downarrow0}g(x) = \frac{\lim_{x\downarrow0} \frac{d}{d x}\left[ 1-F_-\left( F_+^{-1}(1-x) \right)\right]}{\lim_{x\downarrow0} \ \ 1}
	= \lim_{x\downarrow0} \left[ \frac{-f_-\left( F_+^{-1}(1-x)\right)}{-f_+\left( F_+^{-1}(1-x)\right)} \right]
	= \lim_{t\uparrow M_+}\frac{f_-(t)}{f_+(t)} =k \]
	to yield (b).
	\end{proof}
	
	\item[P3.] 
	\begin{proof}
		Concavity or convexity of the ROC curve implies continuity of the distribution functions, and hence existence of densities.
		From equation (\ref{eq:roccdfs}),
		\begin{equation*}
		\frac{dy_{\mathrm{ROC}}}{dx_{\mathrm{ROC}}} = \frac{-f_+\left( F_-^{-1}(1-x_{\mathrm{ROC}})\right)}{-f_-\left( F_-^{-1}(1-x_{\mathrm{ROC}})\right)} = \frac{f_+(t)}{f_-(t)}, \quad t=F_-^{-1}(1-x_{\mathrm{ROC}}),
		\end{equation*}
		and define $h(t)=f_+(t)/f_-(t)$. If the ROC curve is concave (meaning $\frac{dy_{\mathrm{ROC}}}{dx_{\mathrm{ROC}}}$ is nondecreasing as $x_{\mathrm{ROC}}$ decreases), then $h(t)$ is nondecreasing as $t$ increases, i.e., $h'(t)\geq 0$. Similarly, $h'(t)\leq 0$ if the ROC curve is convex. Now consider the PR curve as defined in ($\ref{eq:p2a}$) and whose monotonicity is determined by the monotonicity of $g(x)$. By ($\ref{eq:p2b}$), the sign of $g'(x)$ equals the sign of 
		\[ c(t)=\frac{1}{h(t)}\left[1-F_+(t)\right] -\left[1-F_-(t)\right], \quad t=F_+^{-1}(1-x). \]
		Moreover, 
		\begin{eqnarray*}
			c'(t) &=& -\frac{h'(t)}{\left[h(t)\right]^2}\left[1-F_+(t)\right] - \frac{f_+(t)}{h(t)} +  f_-(t) \\
			&=& -h'(t)\frac{\left[1-F_+(t)\right]}{\left[h(t)\right]^2} .
		\end{eqnarray*}
		It is helpful the clarify the role of thereshold $t$. The earlier view of $t=F_-^{-1}(1-x_{\mathrm{ROC}})$ allowed us to characterize the shape of function $h(t)$ for changing values of threshold $t$: either $h'(t)\geq0$ or $h'(t)\leq0$ for all $t$, based on concavity or convexity of the ROC curve. For studying monotonicity of the PR curve, however, the threshold must be viewed as $t=F_+^{-1}(1-x_{\mathrm{PR}})$ as used in the expression for $c(t)$ above.
		\begin{itemize}
			\item[(a)] If the ROC curve is concave, then $h'(t)\geq 0$ implies $c'(t)\leq 0$, so $c(t)$ is nonincreasing as $t$ increases. As a result,
			\[ c(t) \geq \lim_{t\uparrow M_+}c(t) = \lim_{t\uparrow M_+} \left\{ \frac{f_-(t)}{f_+(t)}\left[1-F_+(t)\right] \right\} -\left[1-F_-(M_+)\right]. \]
			But $M_+\geq M_-$ implies $1-F_-(M_+)=0$, so 
			\[ c(t) \geq  \lim_{t\uparrow M_+} \left\{ \frac{f_-(t)}{f_+(t)}\left[1-F_+(t)\right] \right\} \geq 0. \]
			Consequently, $g'(x)\geq 0$, and so the PR curve is nonincreasing.
			\item[(b)] If the ROC curve is convex, then $h'(t)\leq 0$ implies $c'(t)\geq 0$, so $c(t)$ is nondecreasing as $t$ increases. As a result,
			\[ c(t) \leq \lim_{t\uparrow M_+}c(t) = \lim_{t\uparrow M_+} \left\{ \frac{f_-(t)}{f_+(t)}\left[1-F_+(t)\right] \right\} -\left[1-F_-(M_+)\right]. \]
			But $M_+< M_-$ implies $1-F_-(M_+)>0$, and $\lim_{t\uparrow M_+} \left\{ \frac{f_-(t)}{f_+(t)}\left[1-F_+(t)\right] \right\}=0$, so $c(t) \leq 0$.
			Consequently, $g'(x)\leq 0$, and so the PR curve is nondecreasing.
		\end{itemize}
			
	\end{proof}
	
	\item[P4.] 
	\begin{proof} 
	For the ROC curve, $F_-(\cdot)=F_+(\cdot)$ applied to (\ref{eq:roccdfs}) yields
	\begin{equation*}
	y_{\mathrm{ROC}}(x_{\mathrm{ROC}}) = 1-F_+\left( F_+^{-1}(1-x_{\mathrm{ROC}}) \right) \leq 1-(1-x_{\mathrm{ROC}}) = x_{\mathrm{ROC}}, \qquad 0< x_{\mathrm{ROC}} < 1,
	\end{equation*}
	by property of generalized inverse distribution functions, with equality when $F_+(\cdot)$ is strictly increasing.
	
	For the PR curve, $F_-(\cdot)=F_+(\cdot)$ applied to (\ref{eq:prcdfs}) yields
	\begin{equation*}
	y_{\mathrm{PR}}(x_{\mathrm{PR}}) = 
	\frac{\pi_+}{\pi_+ + \pi_- \underbrace{ \left[ 1- \underbrace{F_+\left( F_+^{-1}(1-x_{\mathrm{PR}}) \right)}_{\geq (1-x_{\mathrm{PR}})} \right]/x_{\mathrm{PR}} }_{\leq 1} } \geq \pi_+, \qquad 0< x_{\mathrm{PR}} < 1,
	\end{equation*}
	again with equality when $F_+(\cdot)$ is strictly increasing.
	\end{proof} 
	
	\item[P5.] 
	\begin{proof}
		Let us first consider $x_{\mathrm{ROC}}$ for different values of threshold $t$. For $M_+>t>m_+$, $F_-(t)=1$ because $M_-<m_+$, hence $x_{\mathrm{ROC}}=1-F_-(t)=0$. For the same reason, $F_-(t)=1$ for $m_+\geq t>M_-$, so $x_{\mathrm{ROC}}=0$. For other allowable thresholds, $M_-\geq t>m_-$, $x_{\mathrm{ROC}}=1-F_-(t) \geq 0$ and increases to one as $t$ decreases.
		
		Now consider $y_{\mathrm{ROC}}$. For $M_+>t\geq m_+$, $y_{\mathrm{ROC}}=1-F_+(t) \geq 0$ and increases to one as $t$ decreases. Because $F_+(t)=0$ for $t< m_+$, $y_{\mathrm{ROC}}=1-F_+(t)=1$ for $m_+>t>m_-$. This completes the perfect-separation ROC curve. Note that property P1 regards $\lim_{x_{\mathrm{ROC}}\downarrow 0}y_{\mathrm{ROC}}\left( x_{\mathrm{ROC}} \right)=1-F_+(M_-)=1$, thus ignoring the other possible values of $y_{\mathrm{ROC}}$ that correspond to $x_{\mathrm{ROC}}=0$, and that the resulting $y_{\mathrm{ROC}}\left( x_{\mathrm{ROC}} \right)$ becomes a legitimate function. 
		
		Given the  perfect-separation ROC curve, use (\ref{eq:pr-from-roc}) to get
		\[ x_{\mathrm{PR}}=y_{\mathrm{ROC}} = 
		\begin{cases}
		1-F_+(t) & M_+>t> m_+ \\ 1 & m_+\geq t>M_- \\ 1 & M_-\geq t>m_-
		\end{cases} \]
		and
		\[ y_{\mathrm{PR}}=\frac{\pi_+}{\pi_+ + \pi_- \left( x_{\mathrm{ROC}}/y_{\mathrm{ROC}}\right) } = 
		\begin{cases}
		\frac{\pi_+}{\pi_+ + \pi_- \left( 0/\left[ 1-F_+(t) \right] \right) } =1 & M_+>t> m_+ \\ 
		\frac{\pi_+}{\pi_+ + \pi_- \left( 0/1 \right) } =1 & m_+\geq t>M_- \\ 
		\frac{\pi_+}{\pi_+ + \pi_- \left( \left[ 1-F_-(t) \right]/1 \right) }  & M_-\geq t>m_-
		\end{cases}, \]
		which yields the result. Note that property P2 regards $\lim_{x_{\mathrm{PR}}\uparrow1}y_{\mathrm{PR}}\left( x_{\mathrm{PR}} \right) = \frac{\pi_+}{\pi_+ + \pi_- [1-F_-(m_+)]}=1$, thus ignoring the other possible values of $y_{\mathrm{PR}}$ that correspond to $x_{\mathrm{PR}}=1$, and that the resulting $y_{\mathrm{PR}}\left( x_{\mathrm{PR}} \right)$ becomes a legitimate function.
	\end{proof}
		
	\item[P6.] 
	\begin{proof}
		First consider $x_{\mathrm{ROC}}$. For $M_->t\geq m_-$, $x_{\mathrm{ROC}}=1-F_-(t)\geq 0$ and increases to one as $t$ decreases. For $t< m_-$, $F_-(t)=0$ and so $x_{\mathrm{ROC}}=1$.
		
		Now consider $y_{\mathrm{ROC}}$. For $M_->t>M_+$, $y_{\mathrm{ROC}}=1-F_+(t)=0$ because $M_+<m_-$. For $ M_+\geq t>m_+$, $y_{\mathrm{ROC}}=1-F_+(t)\geq 0$ and increases to one as $t$ decreases. This completes the reverse-separation ROC curve.
		
		Given the  reverse-separation ROC curve, 
		\[ x_{\mathrm{PR}}=y_{\mathrm{ROC}} = 
		\begin{cases}
		0 & M_->t>m_- \\
		0 & m_-\geq t>M_+ \\ 
		1-F_+(t) & M_+\geq t>m_+ 
		\end{cases} \]
		and use a slight modification of (\ref{eq:pr-from-roc}) to get
		\[ y_{\mathrm{PR}}=\frac{\pi_+\cdot y_{\mathrm{ROC}}}{\pi_+\cdot y_{\mathrm{ROC}} + \pi_-\cdot  x_{\mathrm{ROC}} } = 
		\begin{cases}
		\frac{\pi_+\cdot 0}{\pi_+\cdot 0 + \pi_- \left[ 1-F_-(t) \right] } =0 & M_->t>m_- \\ 
		\frac{\pi_+\cdot 0}{\pi_+\cdot 0 + \pi_- \cdot 1 } =0 & m_-\geq t>M_+ \\ 
		\frac{\pi_+\cdot \left[ 1-F_+(t) \right]}{\pi_+\cdot \left[ 1-F_+(t) \right] + \pi_- \cdot 1  }  & M_+\geq t>m_+
		\end{cases}, \]
		which yields the reverse-separation PR curve.
		
		Because $\tpr$ is a probability, $y_{\mathrm{ROC}}(x_{\mathrm{ROC}})$ is clearly bounded below by zero, and this lower bound is achieved by the reverse-separation ROC curve.
		
		As a probability, precision (i.e., $y_{\mathrm{PR}}$) is clearly bounded below by zero, but there is a much tighter bound. By properties of distribution functions, $0\leq 1-F_-\left(F_+^{-1}(1-x_{\mathrm{PR}})\right)\leq 1$, so
		\[ \pi_+\cdot x_{\mathrm{PR}} + \pi_-\left[ 1-F_-\left(F_+^{-1}(1-x_{\mathrm{PR}})\right) \right] \leq \pi_+\cdot x_{\mathrm{PR}} + \pi_- .\]
		Hence, with a slight modification to (\ref{eq:prcdfs}),
		\[ y_{\mathrm{PR}}(x_{\mathrm{PR}}) = \frac{\pi_+\cdot x_{\mathrm{PR}}}{\pi_+\cdot x_{\mathrm{PR}} + \pi_-\left[ 1-F_-\left(F_+^{-1}(1-x_{\mathrm{PR}})\right) \right]} \geq \frac{\pi_+\cdot x_{\mathrm{PR}}}{\pi_+\cdot x_{\mathrm{PR}} + \pi_-}.  \]
		Furthermore, this lower bound is achieved by the reverse-separation PR curve as demonstrated above.
	\end{proof}
	
	\item[P7.] 
	\begin{proof}
		First consider an ROC curve based on the original score $S$:
		\[ x_{\mathrm{ROC}} = \Pr(S>t|-) , \qquad y_{\mathrm{ROC}} = \Pr(S>t|+) , \qquad m_-<t<M_-. 	 \]
		Now consider an increasing function $h(\cdot)$ with inverse $h^{-1}(\cdot)$. Then the ROC curve based on the transformed score $h(S)$, using thresholds $r$ such that $h(m_-)<r<h(M_+)$, is
		\[ \begin{array}{rllll}
		x_{h,\mathrm{ROC}} & = \Pr(h(S)>r|-) & = \Pr(S>h^{-1}(r)|-) & \stackrel{t=h^{-1}(r)}{=}\Pr(S>t|-) &=x_{\mathrm{ROC}} \\
		y_{h,\mathrm{ROC}} & = \Pr(h(S)>r|+) & = \Pr(S>h^{-1}(r)|+) & \stackrel{t=h^{-1}(r)}{=}\Pr(S>t|+) &=y_{\mathrm{ROC}} .		
		\end{array} \]
		Hence, the transformed scores lead to the same ROC curve as the original scores.
		
		Similarly,
		\[\begin{array}{rlll}
		x_{h,\mathrm{PR}} & = y_{h,\mathrm{ROC}} & = y_{\mathrm{ROC}} & = x_{\mathrm{PR}} \\
		y_{h,\mathrm{PR}} & = \frac{\pi_+}{\pi_++\pi_- \cdot x_{h,\mathrm{ROC}}/y_{h,\mathrm{ROC}}} & = \frac{\pi_+}{\pi_++\pi_- \cdot x_{\mathrm{ROC}}/y_{\mathrm{ROC}}} & = y_{\mathrm{PR}}
		\end{array}. \]
		Hence, the transformed scores lead to the same PR curve as the original scores.
	\end{proof}
	
\end{itemize}

\vskip 0.2in
\bibliography{PrecisionRecall}

\end{document}